\documentclass[nohyperref]{article}

\usepackage{microtype}
\usepackage{graphicx}
\usepackage{subfigure}
\usepackage{booktabs} %
\usepackage{multirow}
\usepackage{bm}
\usepackage{bbm}
\usepackage[page]{appendix}
\usepackage[table,xcdraw]{xcolor}
\usepackage{xcolor, color, soul}
\usepackage{tcolorbox}
\usepackage{hyperref}
\usepackage[normalem]{ulem}
\usepackage{listings}

\usepackage[accepted]{icml2023}

\usepackage{amsmath}
\usepackage{amssymb}
\usepackage{mathtools}
\usepackage{amsthm}

\usepackage[capitalize,noabbrev]{cleveref}

\definecolor{codegreen}{rgb}{0,0.6,0}
\definecolor{codegray}{rgb}{0.5,0.5,0.5}
\definecolor{codepurple}{rgb}{0.58,0,0.82}
\definecolor{backcolour}{rgb}{0.95,0.95,0.92}

\lstdefinestyle{mystyle}{
    backgroundcolor=\color{backcolour},   
    commentstyle=\color{codegreen},
    keywordstyle=\color{magenta},
    numberstyle=\tiny\color{codegray},
    stringstyle=\color{codepurple},
    basicstyle=\ttfamily\footnotesize,
    breakatwhitespace=false,         
    breaklines=true,                 
    captionpos=b,                    
    keepspaces=true,                 
    numbersep=5pt,                  
    showspaces=false,                
    showstringspaces=false,
    showtabs=false,                  
    tabsize=2
}
\theoremstyle{plain}

\theoremstyle{definition}

\theoremstyle{remark}

\newcommand{\hypothesisincorrect}[2]{%
\begin{tcolorbox}[colback=gray!10!white,leftrule=2.5mm,size=title]
\textbf{#1}: #2
\end{tcolorbox}
\vspace{-0.1cm}%
}

\newcommand{\hypothesiscorrect}[2]{%
\begin{tcolorbox}[colback=gray!10!white,leftrule=2.5mm,size=title]
\textbf{#1}: #2
\end{tcolorbox}
\vspace{-0.1cm}%
}

\newcommand{\finding}[2]{%
\begin{tcolorbox}[colback=yellow!10!white,leftrule=2.5mm,size=title]
\textbf{#1}: #2
\end{tcolorbox}
\vspace{-0.1cm}%
}

\usepackage[textsize=tiny]{todonotes}

\newcommand{\orig}[1]{{\color[HTML]{AF1E1E} #1}}
\newcommand{\ind}[1]{{\color[HTML]{00D59D} #1}}
\newcommand{\labs}[1]{{\color[HTML]{3492C7} #1}}
\newcommand{\nearopt}[1]{{\color[HTML]{BF9907} #1}}
\newcommand{\gray}[1]{{\color{gray} #1}}

\newcommand{\R}{\mathbb{R}}

\sethlcolor{lightgray}

\icmltitlerunning{When does Privileged Information Explain Away Label Noise?}

\begin{document}

\twocolumn[
\icmltitle{When does Privileged Information Explain Away Label Noise?}

\icmlsetsymbol{equal}{*}

\begin{icmlauthorlist}
\icmlauthor{Guillermo Ortiz-Jimenez}{equal,epfl}
\icmlauthor{Mark Collier}{equal,google}
\icmlauthor{Anant Nawalgaria}{google}
\icmlauthor{Alexander D'Amour}{google}
\icmlauthor{Jesse Berent}{google}
\icmlauthor{Rodolphe Jenatton}{google}
\icmlauthor{Effrosyni Kokiopoulou}{google}
\end{icmlauthorlist}

\icmlaffiliation{epfl}{Ecole Polytechnique F\'ed\'erale de Lausanne (EPFL). Work done during an internship at Google.}
\icmlaffiliation{google}{Google Research}

\icmlcorrespondingauthor{Guillermo Ortiz-Jimenez}{guillermo.ortizjimenez@epfl.ch}
\icmlcorrespondingauthor{Mark Collier}{mark.collier@google.com}

\icmlkeywords{Machine Learning, ICML}

\vskip 0.3in
]

\printAffiliationsAndNotice{\icmlEqualContribution} %

\begin{abstract}
Leveraging \emph{privileged information} (PI), or features available during training but not at test time, has recently been shown to be an effective method for addressing label noise. However, the reasons for its effectiveness are not well understood. In this study, we investigate the role played by different properties of the PI in explaining away label noise. Through experiments on multiple datasets with real PI (CIFAR-N/H) and a new large-scale benchmark ImageNet-PI, we find that PI is most helpful when it allows networks to easily distinguish clean from mislabeled data, while enabling a learning shortcut to memorize the mislabeled examples. Interestingly, when PI becomes too predictive of the target label, PI methods often perform worse than their no-PI baselines. Based on these findings, we propose several enhancements to the state-of-the-art PI methods and demonstrate the potential of PI as a means of tackling label noise. Finally, we show how we can easily combine the resulting PI approaches with existing no-PI techniques designed to deal with label noise.\looseness=-1
\end{abstract}

\section{Introduction}
Label noise, or incorrect labels in training data, is a pervasive problem in machine learning that is becoming increasingly common as we train larger models on more weakly annotated data. Human annotators are often the source of this noise, assigning incorrect labels to certain examples~\citep{snow-etal-2008-cheap,sheng_2008}, e.g., when the class categories are too fine-grained. Incorrect labeling can also come from using other models to provide proxy labels~\citep{attention_conditioned_masking_consistency} or scraping the web~\citep{radford2021learning}. However, the standard approach in supervised learning is to ignore this issue and treat all labels as correct, leading to significant drops in model performance as the models tend to memorize the noisy labels and degrade the learned representations~\citep{ZhangBengioHardtRechtVinyals.17}.\looseness=-1

Recently, some studies have proposed to mitigate the effect of label noise by leveraging \emph{privileged information} (PI)~\citep{lupi_vapnik, tram} i.e., features available at training time but not at test time. Examples of PI are features describing the human annotator that provided a given label, such as the annotator ID, the amount of time needed to provide the label, the experience of the annotator, etc. While several PI methods have shown promising gains in mitigating the effects of label noise~\citep{distill_pi, gaussian_dropout, tram}, the reasons behind their success are not fully understood. Moreover, the fact that existing results have been provided in heterogeneous settings makes comparisons and conclusions difficult to be drawn.
\looseness=-1

\begin{figure*}[t]
    \centering
    \includegraphics[width=\textwidth]{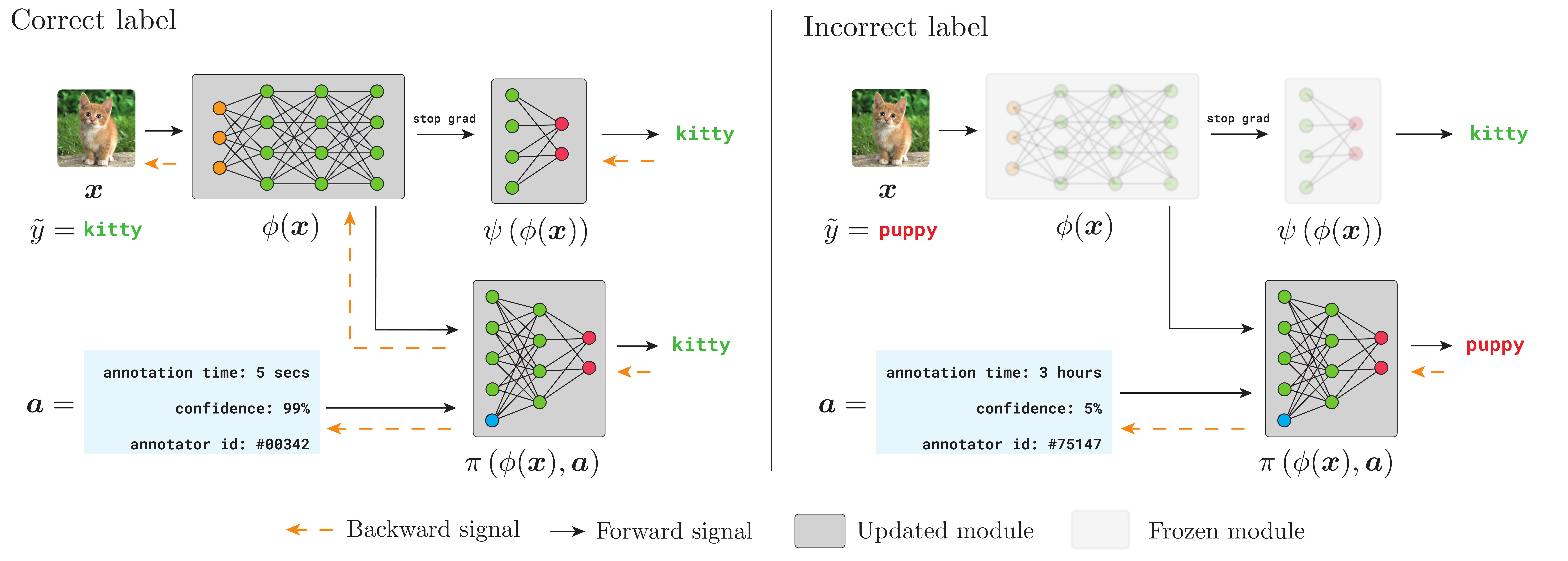}
    \caption{Conceptual illustration of ideal signal propagation while training a privileged information method such as TRAM~\citep{tram} with noisy labels. Having access to PI allows a network to use a learning shortcut to memorize the mislabeled examples using only PI. This protects the extraction of features from the actual data, which are only refined using the correctly labeled examples.}
    \label{fig:illustration}
\end{figure*}

In this work, we aim to standardize the evaluation of PI and conduct a large-scale study on the role of PI in explaining away label noise. We examine the performance of several PI methods on different noisy datasets and analyze their behavior based on the predictive properties of the available PI. Interestingly, we find that when the PI is too predictive of the target label, the performance of most PI methods significantly degrades below their no-PI baselines as the models fail to learn the associations between the non-privileged input features and the targets. Conversely, we discover that the strongest form of PI exhibits two main properties: (i) it allows the network to easily separate clean and mislabeled examples, and (ii) it enables an easier ``learning shortcut" (in a sense to be clarified later) to overfit to the mislabeled examples (see~\cref{fig:illustration}). When both these properties are present, the performance of PI methods significantly exceeds their no-PI counterparts by becoming more robust to label noise.\looseness=-1

Overall, we observe that using PI during training can enable models to discover shortcuts that can prevent learning the relationship between features and labels~\citep{shortcut_geirhos, alex_underspecification}. On the one hand, these \emph{PI-enabled shortcuts}
can have a positive effect when the relationship being ignored only concerns incorrect labels, which thus prevents the learned feature representations from being contaminated by label noise. 
On the other hand, they can have a detrimental effect when the clean labels are also affected, which prevents the model from learning the correct association between features and targets.
Shortcuts are key to understanding the role of PI in mitigating label noise and are directly linked to deep learning dynamics~\citep{ZhangBengioHardtRechtVinyals.17}.\looseness=-1

When focusing on the dynamics of the strongest PI methods, we show that using PI allows training larger models on datasets with a higher level of noise. PI can counteract the negative effect of memorizing incorrect associations between features and incorrect labels as it enables a shortcut that primarily affects the mislabeled examples. We use these new insights to improve current state-of-the-art PI algorithms.\looseness=-1

Overall, the main contributions of our work are:
\begin{itemize}
\vspace{-0.7em}
    \item We present the first large-scale study on the role of PI on supervised noisy datasets of various types.\looseness=-1
    \item We release ImageNet-PI, the largest available testbed for experimenting with PI and label noise.
    \item We find that effective PI enables learning shortcuts only on mislabeled data greatly benefitting performance.\looseness=-1
    \item We improve a wide range of PI methods using simple improvements, and demonstrate cumulative gains with other state-of-the-art noisy labels methods.\looseness=-1
\end{itemize}

We believe our findings can have a significant impact on future research about both label noise and PI. They indeed not only inform us about the desired properties of the ideal PI (which can help design and collect PI features in practice)
but also provide practical insights for improving existing
methods. Formally capturing our empirical results is another promising direction for future research.\looseness=-1

\section{Methodology}

Our large-scale experiments provide a comprehensive analysis of the usefulness of PI in the presence of label noise. Previous results have been provided in heterogeneous settings with testing performed on different datasets and various types of PI, making well-aligned comparisons difficult. We aim at standardizing these comparisons, making the unification of all the settings of our  experiments part of our core contribution. Our code can be found at \url{https://github.com/google/uncertainty-baselines}. In what follows, we briefly describe the datasets and baselines used in our study.\looseness=-1

\subsection{Datasets}\label{sec:datasets}

In this work, we address the supervised learning setting with PI and label noise as described in \citet{tram}. Our training data consists of triplets $(\bm x, \tilde{y}; \bm{a})$, where $\bm x\in\R^d$ is a set of input features, $\tilde{y}\in\{1, \dots, K\}$ is a noisy target label (assuming $K$ classes), and $\bm a\in\R^{p}$ is a vector of PI features. In this work, we mainly focus in the case when these PI features are related to the annotation process, as this is a common
source of label noise~\citep{snow-etal-2008-cheap, sheng_2008}. This PI may include information about the annotator, such as their ID or experience; or about the process itself, such as the annotation duration or confidence. At test time, we do not have access to any PI and evaluate our models based only on clean $(\bm x, y)$ pairs from the data distribution.\looseness=-1

We use relabelled versions of standard image recognition datasets which provide various forms of PI about the annotation process in our experiments. These datasets allow us to access both clean ($y$) and noisy labels ($\tilde{y}$), but we only use the noisy labels for training and hyperparameter selection (see details in \cref{ap:early_stopping} for a discussion about the effect of noisy labels at this step). The clean labels are only used for evaluation. The datasets we use offer a range of training conditions, including differing numbers of samples and classes, levels of noise, types of PI, image sizes, and annotation processes, making our findings widely applicable. \looseness=-1

Some of these datasets provide multiple annotations per example. Nonetheless, to create a unified benchmark we only sample one label per example for datasets that provide multiple labels. So that we can control the noise level and examine its impact on the performance of PI methods, we create high and low noise versions of each dataset, when possible. We follow the terminology of \citet{cifarn} by naming the low noise version the ``uniform'' version, which selects one of the available labels uniformly at random, and the high noise version, the ``worst'' version, which always selects an incorrect label if available. The ``worst'' version is by design more noisy than the ``uniform'' one.\looseness=-1

\paragraph{CIFAR-10/100N.} A relabelled version of the CIFAR-10/100 datasets~\citep{krizhevskyLearningMultipleLayers2009} that includes multiple annotations per image~\citep{cifarn}. The raw data includes information about the annotation process, such as annotation times and annotator IDs, but this information is not given at the example level. Instead, it is provided as averages over batches of examples, resulting in coarse-grained PI. We will show that the PI baselines perform poorly on this dataset. The ``uniform'' version of CIFAR-10N agrees $82.6\%$ of the time with the clean labels, and the ``worst'' version $59.8\%$. CIFAR-100N agrees $59.8\%$ with the clean labels. For reference, training our base architectures without label noise and without PI achieves test accuracies of $93.5\%$ and $77.9\%$ on CIFAR-10 and CIFAR-100, respectively. \looseness=-1

\paragraph{CIFAR-10H.} An alternative human-relabelled version of CIFAR-10, where the new labels are provided only on the test set~\citep{cifarh}. As in \citet{tram}, when we train on CIFAR-10H, we evaluate the performance of the models on the original CIFAR-10 \emph{training set} (since CIFAR-10H relabels only the validation set). Contrary to the CIFAR-N datasets, CIFAR-10H contains rich PI at the example-level, with high-quality metadata about the annotation process. The ``uniform'' version agrees $95.1\%$ of the time with the clean labels, and the ``worst'' $35.4\%$. For reference, training our base architecture without label noise and without PI achieves a test accuracy of $88.4\%$ on CIFAR-10H.\looseness=-1

\paragraph{ImageNet-PI.} Inspired by \citet{tram}, a relabeled version of ImageNet~\citep{dengImageNetLargescaleHierarchical2009} 
in which the labels are provided by a set of pre-trained deep neural networks with different architectures. During the relabeling process, we sample a random label from a temperature-scaled predictive distribution of each model on each example. This leads to label noise that is asymmetrical and feature-dependent. Technical details of the relabeling process and temperature-scaling can be found in \cref{ap:imagenetpi}. The PI of the dataset
comes from the confidences of the models on the sampled labels, the parameter counts of the models, and the models' test accuracies on the clean test distribution. These PI features serve as a good proxy for the expected reliability of each model. The ImageNet-PI high-noise version that
we use agrees $16.2\%$ of the time with the clean labels and the low-noise version $51.9\%$. For reference, training our base architecture without label noise and without PI achieves a test accuracy of $76.2\%$ on ImageNet. As a contribution of this work, we open-source ImageNet-PI (with different amounts of label noise) 
to encourage further research on PI and label noise at a scale larger than possible today with CIFAR-N/H. The data is publicly available at \url{https://github.com/google-research-datasets/imagenet_pi}.
\looseness=-1

\subsection{PI Algorithms}\label{sec:methods}

We study the performance of four representative approaches that exploit PI. They all have been shown to be effective at mitigating the effect of label noise \citep{tram}:\looseness=-1

\paragraph{no-PI.} A standard supervised learning baseline that minimizes the cross-entropy loss on the noisy labels to approximate $p(\tilde{y}|\bm{x})$ without access to PI.
\paragraph{Distillation \citep{distill_pi}.} A knowldege distillation method in which a teacher model is first trained using standard maximum likelihood estimation with access to PI to approximate $p(\tilde{y}|\bm{x}, \bm{a})$. A student model with the same architecture is later trained to match the output of the teacher without access to the PI. We also provide results for a standard self-distillation baseline in which the teacher model does not have access to PI~\citep{standard_distillation}.
\paragraph{TRAM \citep{tram}.} Method based on a two-headed model in which one head has access to PI and the other one not. At training time, a common feature representation $\phi(\bm x)$ is fed to two classification heads $\pi(\phi(\bm x), \bm a)$ (``PI head'') and $\psi(\phi(\bm x))$ (``no-PI head'') to jointly solve\looseness=-1
\begin{equation}\label{eq:tram}
    \min_{\phi,\pi,\psi}\mathbb{E}_{(\bm x, \bm a, \tilde{y})}\left[\ell(\pi(\phi(\bm x), \bm a), \tilde{y})+\ell(\psi(\phi(\bm x)), \tilde{y})\right].
\end{equation}
Importantly, during training, the no-PI feature extractor $\phi$ is updated using \emph{only} the gradients coming from the PI head.
At test time, only the no-PI head
is used for prediction.
\paragraph{Approximate Full Marginalization \citep{tram}.}
A neural network is first trained using maximum likelihood estimation with access to PI to approximate $p(\tilde{y}|\bm{x}, \bm{a})$. During inference, a Monte-Carlo estimation is used to approximate the
marginal\looseness=-1
\begin{equation*}
    p(\tilde{y}|\bm{x})=\int p(\tilde{y}|\bm{x},\bm{a})p(\bm{a}|\bm{x})\;\mathrm{d}\bm a
\end{equation*}
typically further assuming the independence $p(\bm{a}|\bm{x})\approx p(\bm a)$. Note that this process increases the memory and computational costs during inference as it requires computing the output of the network for each of the different sampled values of $\bm a$ (in practice, in the order of $1,000$ extra values).

All the methods use the same underlying network architecture with minimal changes to accommodate their specific requirements, like \citet{tram}. 
In particular, during inference, all methods use exactly the
same network, except for the approximate full marginalization (AFM) baseline, which has additional parameters to deal with the sampled PI.\looseness=-1

In all experiments, we use the same protocol for training with noisy labels and evaluating on a clean test set. As early stopping is a strong baseline against label noise~\citep{understanding_early}, we always report values of test accuracy at the end of the epoch that achieves the best performance on a held-out validation percentage of the \textit{noisy} training set. We reproduce our results without early stopping in \cref{ap:early_stopping}. Unless otherwise specified, we conduct a grid search to tune the most important hyperparameters of each method for each experiment, and report the mean test accuracy and standard deviation over five runs. Further details on the experimental setup can be found in \cref{ap:experimental_details}.

\section{When is PI helpful?}

\begin{table*}[t]
\centering

\caption{Test accuracy of several methods trained using different features as PI (baselines in \gray{gray} and \emph{italics} do not use PI). Here, \orig{Original} denotes the standard PI of the dataset, \ind{Indicator} a binary signal that separates clean from noisy examples, \labs{Labels} the one-hot encoded labels, and \nearopt{Near-optimal} a synthetic feature 
that gives
the annotator label to those examples that are miss-annotated and a zero-vector otherwise. \textbf{Bold numbers} represent significant maximum values across PI features where significance means $\text{p-value}<0.05$.}
\label{tab:pi_types}
\vspace{1em}

\begin{tabular}{@{}clcccccc@{}}
\toprule
\multicolumn{1}{l}{} & \multicolumn{1}{c}{} & \orig{Original}                     & \ind{Indicator}              & \labs{Labels}                       & \nearopt{Near-optimal}                \\ \midrule
& \emph{no-PI}          & \gray{$55.0\scriptstyle{\pm 1.5}$} & \gray{$55.0\scriptstyle{\pm 1.5}$} &  \gray{$55.0\scriptstyle{\pm 1.5}$} & \gray{$55.0\scriptstyle{\pm 1.5}$} \\
 CIFAR-10H           & \emph{Distillation (no-PI)}   & \gray{$47.9\scriptstyle{\pm 0.0}$} & \gray{$47.9\scriptstyle{\pm 0.0}$} &  \gray{$47.9\scriptstyle{\pm 0.0}$} & \gray{$47.9\scriptstyle{\pm 0.0}$} \\
 (worst)           & TRAM          & \orig{$64.9\scriptstyle{\pm 0.8}$} & \ind{$63.3\scriptstyle{\pm 1.1}$} &  \labs{$38.3\scriptstyle{\pm 0.2}$} & \nearopt{$\bm{67.8\scriptstyle{\pm 0.2}}$} \\
                     & Approximate FM            & \orig{$64.0\scriptstyle{\pm 0.6}$}   & \ind{$66.7\scriptstyle{\pm 2.1}$}   &  \labs{$29.5\scriptstyle{\pm 0.5}$}   & \nearopt{$\bm{74.4\scriptstyle{\pm 0.1}}$}   \\
                     & Distillation (PI)       & \orig{$45.4\scriptstyle{\pm 0.8}$} & \ind{$\bm{49.9\scriptstyle{\pm 0.7}}$} &  \labs{$44.5\scriptstyle{\pm 0.1}$} & \nearopt{$48.2\scriptstyle{\pm 0.9}$} \\ \midrule
                     & \emph{no-PI}          & \gray{$80.6\scriptstyle{\pm 0.2}$} & \gray{$80.6\scriptstyle{\pm 0.2}$} &  \gray{$80.6\scriptstyle{\pm 0.2}$} & \gray{$80.6\scriptstyle{\pm 0.2}$} \\
CIFAR-10N            & \emph{Distillation (no-PI)}   & \gray{$80.4\scriptstyle{\pm 0.0}$} & \gray{$80.4\scriptstyle{\pm 0.0}$} &  \gray{$80.4\scriptstyle{\pm 0.0}$} & \gray{$80.4\scriptstyle{\pm 0.0}$} \\
(worst)            & TRAM          & \orig{$80.5\scriptstyle{\pm 0.5}$} & \ind{$87.9\scriptstyle{\pm 0.4}$} &  \labs{$48.9\scriptstyle{\pm 0.2}$} & \nearopt{$\bm{89.3\scriptstyle{\pm 0.3}}$} \\
                     & Approximate FM            & \orig{$82.0\scriptstyle{\pm 0.3}$}   & \ind{$91.2\scriptstyle{\pm 0.3}$}   &  \labs{$22.6\scriptstyle{\pm 0.2}$}   & \nearopt{$\bm{92.0\scriptstyle{\pm 0.1}}$}   \\
                     & Distillation (PI)       & \orig{$80.2\scriptstyle{\pm 0.3}$} & \ind{$80.1\scriptstyle{\pm 0.3}$} &  \labs{$\bm{80.7\scriptstyle{\pm 0.2}}$} & \nearopt{$80.2\scriptstyle{\pm 0.3}$} \\ \midrule
                     & \emph{no-PI}          & \gray{$60.4\scriptstyle{\pm 0.5}$} & \gray{$60.4\scriptstyle{\pm 0.5}$} &  \gray{$60.4\scriptstyle{\pm 0.5}$} & \gray{$60.4\scriptstyle{\pm 0.5}$} \\
           & \emph{Distillation (no-PI)}   & \gray{$60.6\scriptstyle{\pm 0.2}$} & \gray{$60.6\scriptstyle{\pm 0.2}$} &  \gray{$60.6\scriptstyle{\pm 0.2}$} & \gray{$60.6\scriptstyle{\pm 0.2}$} \\
CIFAR-100N            & TRAM          & \orig{$59.7\scriptstyle{\pm 0.3}$} & \ind{$62.4\scriptstyle{\pm 0.3}$} &  \labs{$34.9\scriptstyle{\pm 0.2}$} & \nearopt{$\bm{67.4\scriptstyle{\pm 0.3}}$} \\
                     & Approximate FM            & \orig{$60.0\scriptstyle{\pm 0.2}$}   & \ind{$66.4\scriptstyle{\pm 0.2}$}   &  \labs{$20.1\scriptstyle{\pm 0.3}$}   & \nearopt{$\bm{70.2\scriptstyle{\pm 0.1}}$}   \\
                     & Distillation (PI)       & \orig{$61.1\scriptstyle{\pm 0.2}$} & \ind{$\bm{61.8\scriptstyle{\pm 0.3}}$} &  \labs{$60.5\scriptstyle{\pm 0.2}$} & \nearopt{$\bm{61.5\scriptstyle{\pm 0.3}}$} \\ \midrule
                     & \emph{no-PI}          & \gray{$47.7\scriptstyle{\pm 0.8}$} & \gray{$47.7\scriptstyle{\pm 0.8}$} &  \gray{$47.7\scriptstyle{\pm 0.8}$} & \gray{$47.7\scriptstyle{\pm 0.8}$} \\
ImageNet-PI          & \emph{Distillation (no-PI)}   & \gray{$50.2\scriptstyle{\pm 0.8}$} & \gray{$50.2\scriptstyle{\pm 0.8}$} &  \gray{$50.2\scriptstyle{\pm 0.8}$} & \gray{$50.2\scriptstyle{\pm 0.8}$} \\
(high-noise)          & TRAM          & \orig{$53.3\scriptstyle{\pm 0.5}$} & \ind{$53.6\scriptstyle{\pm 0.5}$} &  \labs{$41.0\scriptstyle{\pm 0.7}$} & \nearopt{$\bm{56.5\scriptstyle{\pm 0.3}}$} \\
                     & Approximate FM            & \orig{$55.6\scriptstyle{\pm 0.3}$}   & \ind{$55.3\scriptstyle{\pm 0.6}$}   &  \labs{$0.8\scriptstyle{\pm 0.2}$}   & \nearopt{$\bm{58.3\scriptstyle{\pm 0.1}}$}   \\
                     & Distillation (PI)       & \orig{$\bm{51.0\scriptstyle{\pm 0.4}}$} & \ind{$\bm{50.6\scriptstyle{\pm 0.2}}$} &  \labs{$39.0\scriptstyle{\pm 4.6}$} & \nearopt{$27.5\scriptstyle{\pm 22.7}$} \\ \bottomrule
\end{tabular}
\end{table*}

\Cref{tab:pi_types} (\orig{Original}) shows the performance of the different PI algorithms on our collection of noisy datasets\footnote{We present results for high-noise in the main text. A reproduction of \cref{tab:pi_types} with low-noise can be found in \cref{ap:low_noise}.\looseness=-1}, where we see that leveraging PI does not always yield big gains in performance. Indeed, while TRAM and AFM substantially improve upon the
no-PI baseline on CIFAR-10H and ImageNet-PI, they do not perform much better on CIFAR-10N and CIFAR-100N. Moreover, we observe little gains of Distillation (PI) over the vanilla self-distillation baseline.

The performance disparities of the same algorithms on datasets where the main source of variation is the available PI, i.e., CIFAR-10N \emph{vs.}~CIFAR-10H, highlights that leveraging PI is not always helpful. In fact, depending on the predictive properties of the PI and the noise distribution, we report very different results. This begs the questions: i) ``\emph{what makes PI effective for these algorithms?}'' and ii) ``\emph{how do they exploit PI to explain away label noise?}''.

To answer these question, we perform a series of controlled experiments in which we train our PI methods
using different PI features (including both real and synthetic ones). By doing so
our objective is
to identify the main mechanisms that lead to the top performance of these algorithms.

\begin{figure*}[t]
    \centering
    \includegraphics[width=\textwidth]{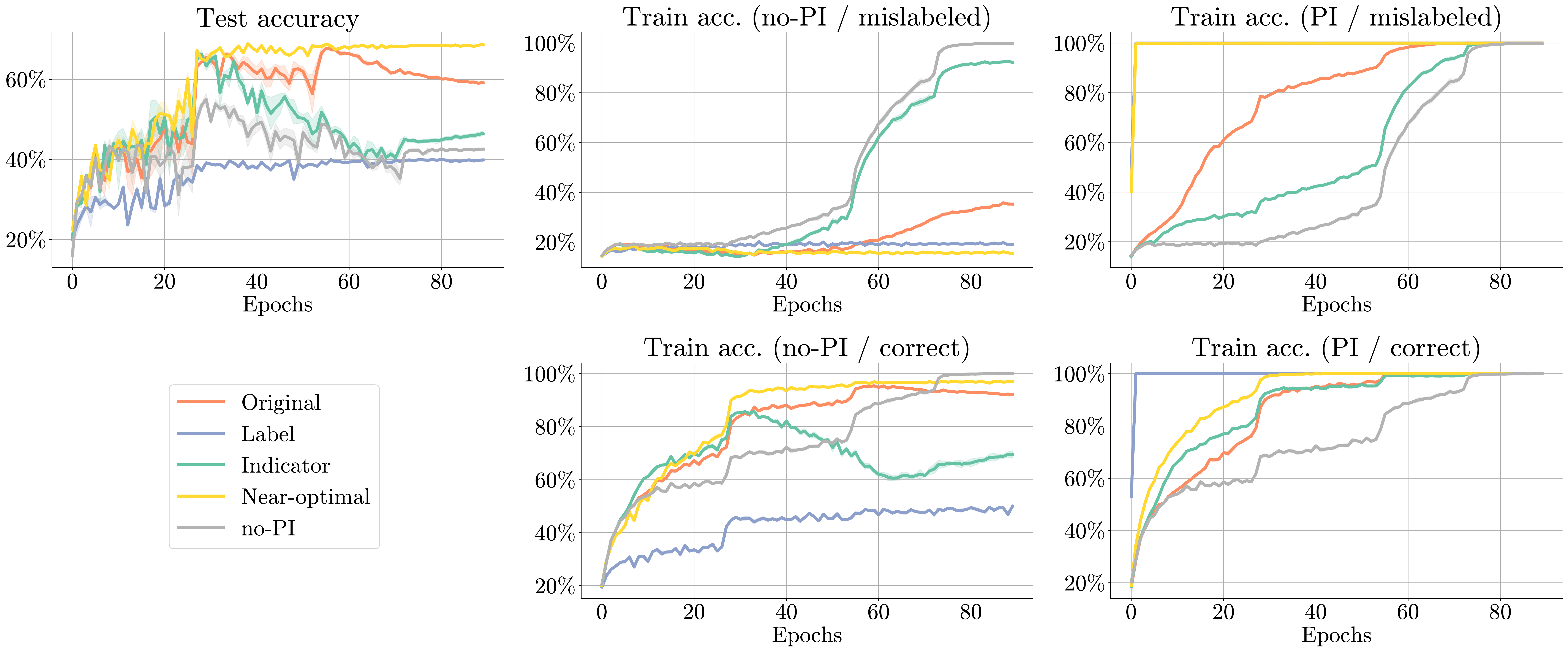}
    \vspace{-2em}
    \caption{Dynamics of TRAM on CIFAR-10H with different PIs. \textbf{Top left}: Test accuracy. \textbf{Top center}: Train accuracy on mislabeled examples evaluated at the no-PI head, \textbf{Top right}: Train accuracy on mislabeled examples evaluated at the PI-head. \textbf{Bottom center}: Train accuracy on clean examples evaluated at the no-PI head. \textbf{Bottom right}: Train accuracy on clean examples evaluated at the PI head.\looseness=-1}
    \label{fig:tram_dynamics}
\end{figure*}

\subsection{Fully predictive PI}
\hypothesisincorrect{Hypothesis}{The PI $\bm{a}$ always complements the information about the labels $\tilde{y}$ contained in $\bm{x}.$}

It is natural to assume that knowing $\bm a$ on top of $\bm x$ can help predict $\tilde{y}$ and thus improve over supervised learning. However, this reasoning is flawed as it forgets that during inference the models cannot exploit $\bm a$. On the contrary, as we will see, if $\bm a$ is very predictive of the target $\tilde{y}$, the test performance can severely degrade.\looseness=-1

We test this hypothesis by retraining the algorithms on the noisy datasets but using $\bm a=\tilde{y}$ instead of the original PI features. That is, having access to fully predictive PI. 

\finding{Finding}{When $\bm{a}$ is fully predictive of $\tilde{y}$, most PI methods perform worse than the no-PI baselines.}

As we can see in \cref{tab:pi_types}~(\labs{Labels}), all the PI baselines greatly suffer in this regime. The reason for this is simple: when the PI is too informative of the target label, then the models are heavily relying on the PI to explain the label and they are discouraged from learning any associations between $\bm x$ and $\tilde{y}$ and do not learn any meaningful feature representations. In this regard, we see how Distillation (PI) achieves roughly the same performance as Distillation (no-PI), while TRAM and AFM achieve very low test accuracies. \looseness=-1

The fact that very predictive PI can hurt the performance of these algorithms highlights a key element of their dynamics: PI can enable learning shortcuts~\citep{alex_underspecification,shortcut_geirhos} that prevent learning certain associations between $\bm x$ and $\tilde y$, possibly by starving the gradient signal that updates $\phi(\bm x)$~\citep{gradient_starvation}. This has practical implications as it discourages blindly appending arbitrarily complex metadata to $\bm a$ during training which could be very predictive of the target label.

\subsection{Noise indicator}
\hypothesisincorrect{Hypothesis}{PI helps because it can separate mislabeled from correct examples.}
We saw that when $\bm a$ is too predictive of $\tilde y$, the PI approaches 
perform poorly. We now turn to an alternative hypothesis of why PI can be beneficial to explain away label noise: 
The PI features can help the network separate the clean from the mislabeled examples. Indeed, the original motivation of using PI to fight label noise in \citet{tram} was that annotator features, e.g., confidences, could act as proxy to identify mislabeled samples.
Intuitively, the main assumption is that if the PI can properly identify the mislabeled examples, then it should act as expert knowledge that would discourage focusing on the hard mislabeled instances, and instead, promote learning only on the correct easy ones~\citep{lupi_vapnik}.
\looseness=-1

Albeit natural, this hypothesis has not been tested before, but can be done using the datasets in this study. Recall that we have access to clean and noisy labels for all the training samples, and thus we can synthesize an indicator signal $\mathbbm{1}(\tilde{y}\neq y)$ that takes a value of $1$ when the clean and noisy labels agree and $0$ otherwise. \Cref{tab:pi_types}~(\ind{Indicator}) shows the results of training using $\bm a =\mathbbm{1}(\tilde{y}\neq y)$. 

\finding{Finding}{Some PI methods perform better with the original PI than with an oracle noise indicator.}

Interestingly, although we see that the performances on the \ind{Indicator} columns are generally higher than on the \orig{Original} one, this is not always the case, and sometimes the \ind{indicator} underperforms or does not significantly improve over the \orig{original} PI (cf. AFM and TRAM on CIFAR10H and ImageNet-PI). This suggests that the PI 
methods
do more than just leveraging the noise indication abilities of the PI. Clearly, if even using an ideal \ind{noise indicator} signal $\mathbbm{1}(\tilde{y}\neq y)$ as PI, the \orig{original} PI can sometimes outperform it, then there must be other information in the PI that the algorithms can exploit to improve performance.

\subsection{Memorization dynamics play a significant role}

Inspecting the training dynamics of the algorithms can help understand the previous results. For example, \cref{fig:tram_dynamics} shows the evolution of test and training accuracies of a TRAM model on CIFAR-10H using different PI features\footnote{Results for others settings can be found in \cref{ap:other_dynamics}.}. The \orig{original} PI leads to better final test accuracy than the \ind{noise indicator}. Meanwhile, models trained using \labs{annotator labels} as PI do not seem to learn anything useful. These differences are explained by the rates at which these models fit the mislabeled and correct samples using each of the TRAM heads.\looseness=-1

Focusing on the training accuracies of the PI-head, \cref{fig:tram_dynamics}~(right column) explains why giving the \labs{labels} as PI hurts the test performance\footnote{The blue line sits behind the yellow line in \cref{fig:tram_dynamics} (top right).}. The \labs{label} model quickly obtains $100\%$ training accuracy on all examples (mislabeled and correct) using the PI head, which in turn slows the training speed of the no-PI head (central column). This happens because the feature extractor is only updated by gradients from the PI head, leading to a lack of meaningful representation of $p(\tilde{y}|\bm x)$ if the model is learning to fit all examples using PI features alone.\looseness=-1

Focusing on the training accuracies of the no-PI head in \cref{fig:tram_dynamics}~(central column), the best models are those that achieve the highest training accuracy on correct examples, while not overfitting to the mislabeled. The difference in test performance of \ind{indicator} and \orig{original} is explained by the \orig{original} model having a harder time overfitting to the mislabeled examples. Interestingly, the \orig{original} model memorizes mislabeled examples faster with the PI head than the \ind{indicator}. It looks as though fitting the training examples fast with the PI head was discouraging the model from fitting the same examples with the no-PI head, i.e., the PI is enabling a learning shortcut to memorize the mislabeled examples with the \orig{original} PI, without using $\bm x$. This might be because the \ind{indicator} signal only takes values in $\{0,1\}$ for all examples, and these are not enough to separate the noisy training set. Indeed, as we will see, having access to PI that can be easily memorized on the mislabeled examples is fundamental to maximize performance.\looseness=-1

\subsection{Near-optimal PI features}

\hypothesiscorrect{Hypothesis}{The optimal PI enables a learning shortcut to memorize \emph{only} the mislabeled labels.}

The experiments using the \labs{annotator labels} as PI are a clear example of a PI-enabled learning shortcut which is very detrimental for the model performance. On the other hand, the dynamics of the \orig{original} models hint that the same shortcut mechanism can also have a positive effect when it only applies to the mislabeled examples. 
To test this hypothesis, we design a new form of PI features, denoted as {\nearopt{near-optimal}} in the tables and plots.
As its name indicates, this PI should allow the models to get very close to their top performance.
The \nearopt{near-optimal} features are designed to exploit the PI shortcut only on the mislabeled examples, allowing the model to learn freely on the correct ones. To that end, the \nearopt{near-optimal} PI features consist of two concatenated values: (i) the indicator signal that says if a given example is mislabeled or not, and (ii) the annotator label \emph{only} if that example is mislabeled. Otherwise an all-zero vector is concatenated with the same dimensionality as the one-hot encoded labels to the indicator signal.\looseness=-1

\finding{Finding}{When a learning shortcut is provided \emph{only} for mislabeled examples, PI methods achieve top performance.\looseness=-1}

The results in \cref{tab:pi_types}~(\nearopt{Near-optimal}) show that those 
PI features significantly outperform all other PI features by a large margin on all datasets when using TRAM or AFM\footnote{\nearopt{Near-optimal} does not always outperform \orig{original} when using Distillation (PI), but note that in general the gains of Distillation (PI) over (no-PI) are much smaller than for TRAM and AFM. In this regard, we leave the objective of finding a near-optimal policy for Distillation (PI) as an open question for future work.\looseness=-1}. Similarly, in \cref{fig:tram_dynamics} we observe that the dynamics of the near-optimal models fully match our expectations. The \nearopt{near-optimal} models train the fastest on the mislabeled examples on the PI head, thus leading to a very slow training speed on mislabeled examples on the no-PI head. Moreover, since the mislabeled examples no longer influence (because their accuracies are already maximal on the PI head) the updates of the feature extraction, then we observe that the performance on the correct examples is much higher.

The same explanation applies to AFM whose dynamics are shown in \cref{ap:other_dynamics}. In this case, the memorization of the mislabeled examples using PI alone also protects the no-PI features. This way, during inference, the PI sampled from the mislabeled examples simply adds a constant noise floor to the predicted probabilities of the incorrect labels. This averaged noise floor is usually much smaller than the probability predicted using the clean features of the no-PI, and thus does not lead to frequent misclassification errors.\looseness=-1

\section{Improving PI algorithms}

In this section, we use the insights of the previous analysis to improve the design of PI methods.
We perform ablation studies on different design parameters of the main PI 
approaches,
identifying simple modifications that significantly boost their performance.
We primarily focus on TRAM and AFM as these methods outperform Distillation (PI) by a large
margin when the PI is helpful (cf. \cref{tab:pi_types}). We provide illustrative results here, and full results 
in
\cref{ap:other_datasets}.\looseness=-1

\subsection{Model size}

We explore how the model size affects performance. In particular, note that the parameter count of all PI algorithms can be split into two parts: 
the feature extractor $\phi$ of the standard features $\bm x$ and the tower $\pi$ that processes the PI; see~Eq.~(\ref{eq:tram}) and \cref{sec:methods}.
We therefore perform an ablation study in which we scale each of these parts of the models separately.\looseness=-1

\begin{figure}[ht]
    \centering
    \includegraphics[width=\columnwidth]{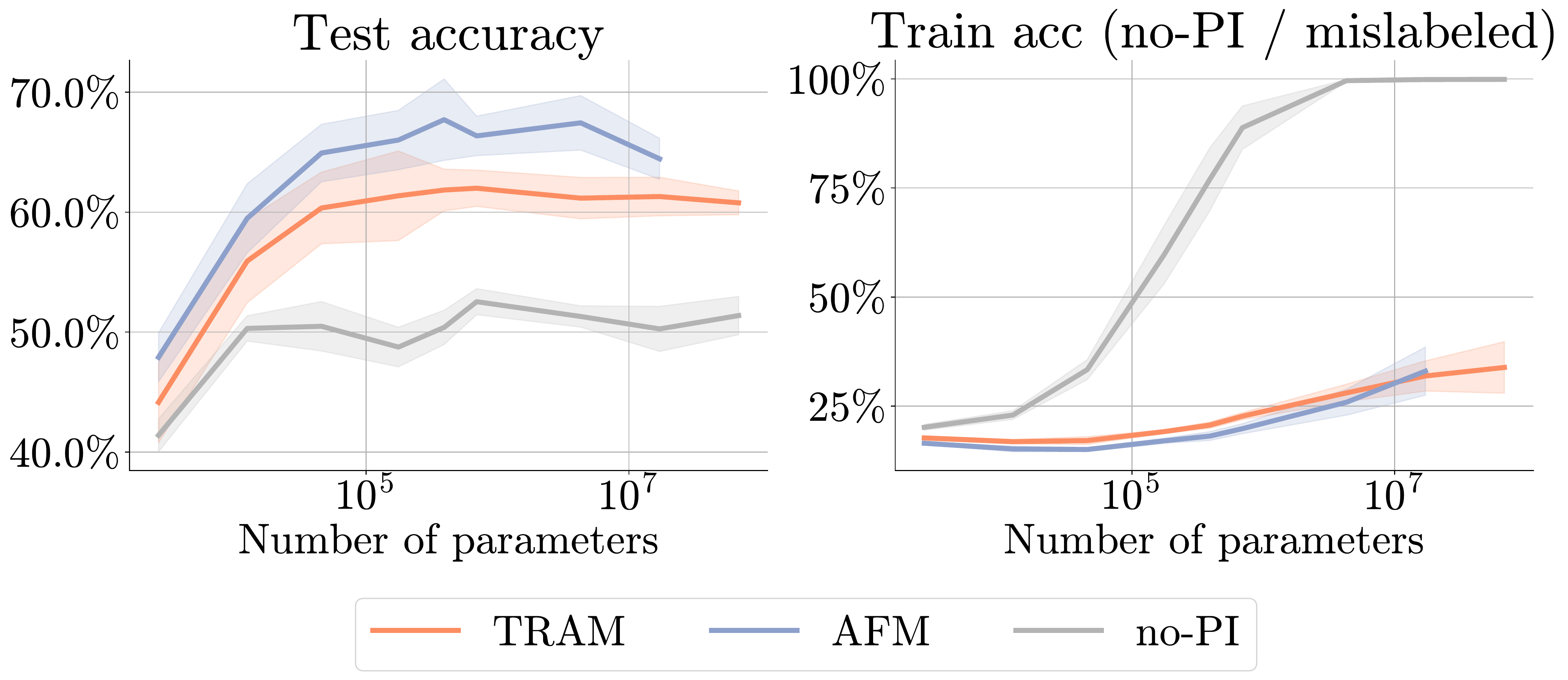}
    \vspace{-2em}
    \caption{Performance of different PI baselines on CIFAR-10H when increasing the parameter count of their feature extractor keeping the PI tower fixed. Larger models suffer from overfitting as they tend to use their larger capacity to overfit to mislabeled examples, discouraging the model from exploiting the PI.\looseness=-1} 
    \label{fig:feature_extractor_size}
\end{figure}

\paragraph{Feature extractor.} \Cref{fig:feature_extractor_size} shows how test accuracy changes as we increase the size of the feature extractor
of the PI approaches.
The performance follows a U-shape, where scaling the model past a certain point harms final performance. 
Indeed, a larger capacity discourages the model from using PI features and causes overfitting to standard features, as shown by the
simultaneous increase in training accuracy on mislabeled examples and decrease in test accuracy.\looseness=-1

\finding{Finding}{Increasing the feature extractor size discourages using the PI as a shortcut.}

\begin{figure}[ht]
    \centering
    \includegraphics[width=\columnwidth]{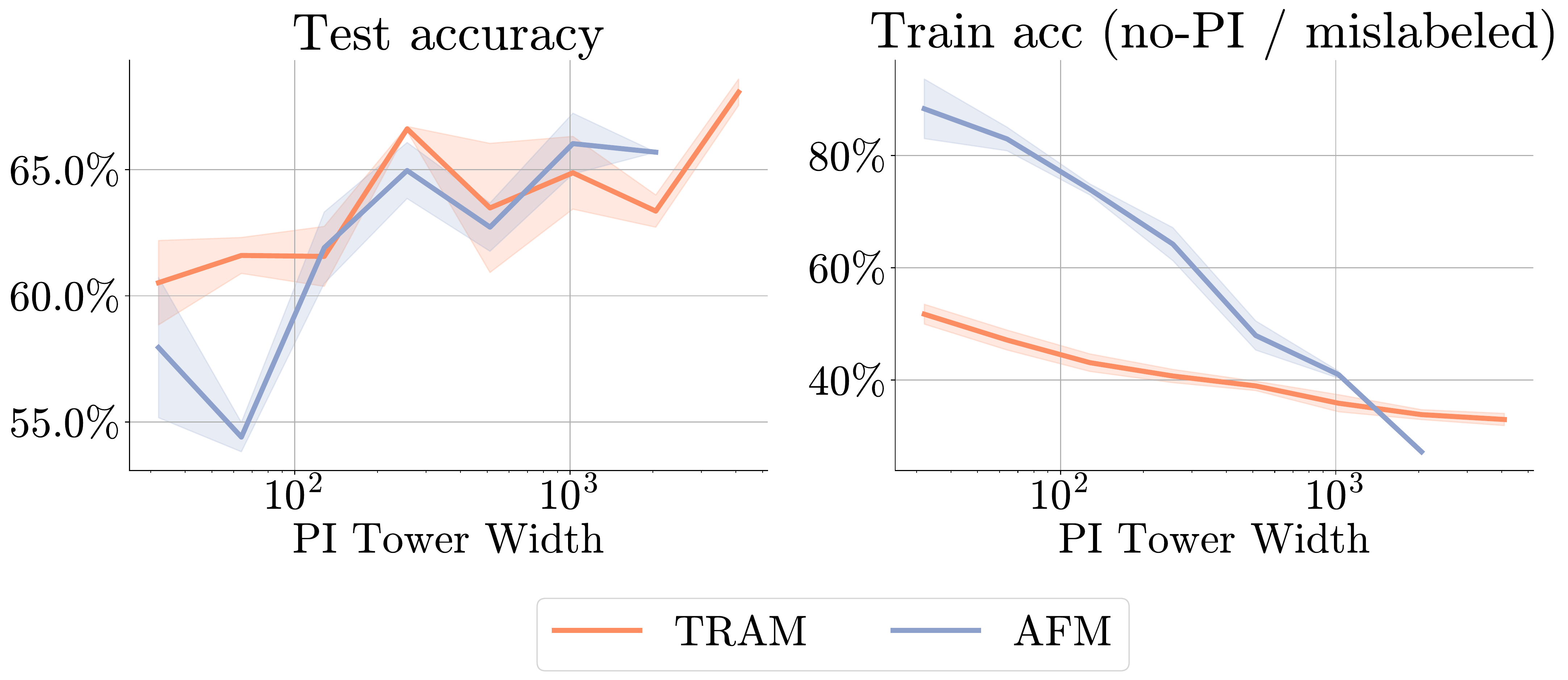}
    \vspace{-2em}
    \caption{Performance of different PI approaches on CIFAR-10H when increasing the PI head size. A larger PI head size incentivizes the model to memorize the mislabeled examples using the PI,
    thus further exploiting PI as a shortcut.\looseness=-1
    }
    \label{fig:pi_head_size}
\end{figure}

\paragraph{PI head size.} \Cref{fig:pi_head_size} shows the results of scaling the size of the PI processing tower while keeping the feature extractor size fixed. We observe how larger PI heads improve performance as they encourage more memorization using PI alone and protect the extraction of the no-PI features. This is illustrated by the decay of the training accuracy of the mislabeled examples on the no-PI head for larger PI heads.
\finding{Finding}{Increasing the capacity of the PI tower
encourages using the PI as a shortcut.}

\subsection{Random PI can enable positive shortcuts} \label{sec:random_pi}

\hypothesiscorrect{Hypothesis}{Random PI that uniquely identifies each example can enable a PI shortcut that protects the model from memorizing incorrect labeles with $\bm x$.}

The \nearopt{near-optimal}, \labs{labels}, and \ind{indicator} signals of \cref{tab:pi_types} are all synthetic PI features that cannot be used in practice, as they rely on the knowledge of which examples are mislabeled and which examples are correct. However, they show that having access to a signal that can be more easily memorized than the standard features $\bm x$ on the mislabeled examples is a good recipe to improve performance. 
This being said,a key property of incorrect labels is that they are, by definition, a mistake. In this sense, fitting an incorrect training label simply amounts to memorizing a specific training example whose features are not predictive of the target label, i.e., 
the features
serve just as an example ID. In fact, any set of features which are different enough for each example could act as such an ID.

\begin{table*}[t!]
\centering
\caption{Performance comparison of no-PI, TRAM, TRAM++, SOP, HET, TRAM+SOP and TRAM+HET on the different PI datasets.}
\label{tab:tram_improvements}
\begin{tabular}{@{}lccc|cc|cc@{}}
\toprule
                          & no-PI & TRAM & TRAM++ & SOP & TRAM+SOP & HET & TRAM+HET         \\ \midrule
CIFAR-10H (worst)         & $55.0\scriptstyle{\pm 1.5}$ & $64.9\scriptstyle{\pm 0.8}$  & $66.8\scriptstyle{\pm 0.3}$ & $59.2 \scriptstyle{\pm 0.8}$ &  $\bm{70.9\scriptstyle{\pm 0.5}}$ & $50.8 \scriptstyle{\pm 1.4}$ & $67.7 \scriptstyle{\pm 0.7}$ \\
CIFAR-10N (worst)         & $80.6\scriptstyle{\pm 0.2}$ & $80.5\scriptstyle{\pm 0.5}$  & $83.9\scriptstyle{\pm 0.2}$ & $87.9 \scriptstyle{\pm 0.2}$ & $\bm{88.5\scriptstyle{\pm 0.3}}$  & $81.9 \scriptstyle{\pm 0.4}$ & $82.0 \scriptstyle{\pm 0.3}$ \\
CIFAR-100N         & $60.4\scriptstyle{\pm 0.5}$ & $59.7\scriptstyle{\pm 0.3}$  & $61.1\scriptstyle{\pm 0.2}$ & $65.3 \scriptstyle{\pm 0.3}$ & $\bm{66.1\scriptstyle{\pm 0.2}}$  & $60.8 \scriptstyle{\pm 0.4}$ & $62.1 \scriptstyle{\pm 0.1}$ \\
ImageNet-PI (high-noise)               & $47.7 \scriptstyle{\pm 0.8}$ & $53.3 \scriptstyle{\pm 0.5}$  & $53.9 \scriptstyle{\pm 0.4}$ & -- & -- & $51.5 \scriptstyle{\pm 0.6}$ & $\bm{55.8 \scriptstyle{\pm 0.3}}$ \\
\bottomrule
\end{tabular}
\end{table*}

\finding{Finding}{Random PI is effective at reducing overfitting to the incorrect labels using $\bm x$.}

We evaluate this hypothesis in \cref{tab:tram_improvements} where we introduce TRAM++: 
a version of TRAM in which the \orig{original} PI features are augmented with a unique random vector for each example (experimental details are provided in \cref{ap:trampp} and results for AFM++ in \cref{ap:afmpp}). As we can see, TRAM++ generally achieves better performance than TRAM alone, with greater improvements in those datasets where overfitting is a bigger issue (i.e., CIFAR).\looseness=-1

\section{Combination with other no-PI techniques}\label{sec:other_techniques}

In this section, we show experimentally that the performance improvements obtained by PI methods on noisy datasets can work symbiotically with other state-of-the-art techniques from the noisy label literature. In particular, we show that TRAM++ can be easily combined with Sparse Over-parameterization (SOP)~\citep{sop} and Heteroscedastic output layers~\citep{het}
while providing cumulative gains with respect to those baselines\footnote{We provide further experiments combining TRAM with label smoothing~\citep{ls} in \cref{ap:label_smoothing}.}.

\subsection{Sparse Over Parameterization (SOP)}

Sparse over-parameterization (SOP) \citep{sop} is a state-of-the-art method which leverages the implicit bias of stochastic gradient descent (SGD) and overparameterization to estimate and correct the noisy label signal, a concept which has proven to work well \citep{Zhao_2022}. It does so by adding two new sets of
$K$-dimensional parameters $\{\bm{u}_i\}_{i=1}^N$ and $\{\bm{v}_i\}_{i=1}^N$,
where $N$ denotes the number of training points, and solving
\begin{equation*}
    \min_{\bm\theta,\{\bm{u}_i, \bm v_i\}_{i=1}^N}\cfrac{1}{N}\sum_{i=1}^N\ell\left(f_{\bm\theta}(\bm x)+\bm u_i\odot\bm u_i -\bm v_i\odot \bm v_i, \tilde{y}\right)
\end{equation*}
using SGD. This specific parameterization biases the solution of SGD towards the recovery of the noise signal $\bm \epsilon_i=\bm u_i\odot\bm u_i -\bm v_i\odot \bm v_i$ that corrupts $y$, i.e., $\tilde{y}_i\approx y_i+\bm\epsilon_i$, implicitly assuming that $\bm\epsilon_i$ is sparse across the dataset.

In this work, we explore whether the combination of TRAM++ with SOP can yield cumulative gains in performance against label noise. In particular, we propose a simple two-step training process to combine them: (i) We first pretrain a neural network using TRAM++ and (ii) we finetune the no-PI side of the network using the SOP loss without stop-gradients. \Cref{tab:tram_improvements} shows the results of this method\footnote{We do not provide results for ImageNet-PI as SOP cannot be easily scaled to such a large dataset.} where we see that, indeed, TRAM+SOP is able to significantly outperform TRAM++ or SOP alone in all datasets%
. More experimental details can be found in \cref{ap:sop}.\looseness=-1

\subsection{Heteroscedastic output layers}

Finally, we further analyze the combination of TRAM with HET, another state-of-the-art no-PI baseline from the noisy label literature that can be scaled 
up to ImageNet scale~\citep{het}.
HET here refers to the use of heteroscedastic output layers to model the 
aleatoric
uncertainty of the predictions without PI. In particular, we apply HET layers to both heads of TRAM++ and follow the same training setup. We call 
the resulting approach
TRAM+HET.\looseness=-1

Our experiments, presented in \cref{tab:tram_improvements}, show that the TRAM+HET model 
outperforms both TRAM++ and HET applied alone
. More experimental details 
about that model combination
can be found in \cref{ap:het}. All in all, these results corroborate our main findings:\looseness=-1

\finding{Finding}{PI methods work symbiotically with other no-PI algorithms from the noisy label literature.}

\section{Related work}

The general framework of learning with privileged information~\citep{lupi_vapnik} has been widely studied in deep learning, with many works exploring different baselines, including loss manipulation~\citep{miml_pi}, distillation~\citep{distill_pi}, or Gaussian dropout~\citep{gaussian_dropout}. This line of work has
mainly focused on the noiseless scenario, conceiving PI as a guiding signal that helps identify easy or hard instances~\citep{svm+}. Similar to our work, \citet{yang2022toward} also studied the role of PI in improving the performance of deep learning methods, but focusing on the task of learning-to-rank using distillation methods in the noiseless setting.\looseness=-1

More recently, \citet{tram} proposed a new perspective on PI, arguing that it can make models more robust to the presence of noise. Their proposed PI approach, referred to as TRAM, led to gains on various experimental settings, with both synthetic and real-world noise. However, their results lacked a detailed analysis of how different sources of PI affect performance.\looseness=-1

Our work takes inspiration from the rich deep-learning theory studying the memorization dynamics of neural networks~\citep{ZhangBengioHardtRechtVinyals.17, rolnick2017deep, toneva_2019, maennel_2020, baldock_2021}. In the no-PI setting, the dynamics of neural networks wherein the incorrect labels tend to be later memorized during training has been heavily exploited by the noisy-label community through techniques such as early-stopping and regularization~\citep{early_learning_reg, understanding_early}. Other works have exploited the intrinsic difference between the learning of clean and mislabeled examples to detect and correct misclassification errors using self-supervision~\citep{google_old_noise, dividemix}, co-teaching~\citep{coteaching}, or regularization~\citep{cores}. Finally, many works have attempted to model the label corruption process by estimating the label transition matrix~\citep{matrix_estimation} or the noisy signal directly in the prediction space~~\citep{sop}. In general, we see this line of research about noisy labels \citep{label_noise_survey} as orthogonal to the use of PI and we have experimentally shown that our PI approach is in fact complementary and can be gracefully combined with such techniques.
\looseness=-1

Some aspects of this work are suggestive of causal reasoning. In particular, \emph{explaining away} is a well-known phenomenon when there are multiple explanations for the value that a particular variable has taken, e.g., whether it is the ground-truth label correctly annotated, or a mistake from an annotator~\citep{pearl}. We do not use causal formalism explicitly in this work, although we see similar learning dynamics at play in our results. PI (often called auxiliary labels) is also used in causally-motivated work on robust ML, although this is usually focused on the distinct problem of handling spurious correlations, rather than overcoming label noise \citep{aahlad,counterfactual_inv,shortcut_removal}. In self-supervised learning, the removal of shortcuts is also a topic of interest~\citep{automatic_shortcut}.\looseness=-1

\section{Conclusions}

In this work, we have presented a systematic study in which we investigate which forms of PI are more effective at explaining away label noise. Doing so we have found that the most helpful PI is the one that allows the networks to separate correct from mislabeled examples in feature space, but also enable an easier learning shortcut to memorize the mislabeled examples. We have also shown that methods which use
appropriate
PI to explain away label noise, can be combined with other state-of-the-art methods to remove noise and achieve cumulative gains. Exploring this direction further is a promising avenue for future work. Our insights show that the use of PI is a promising avenue of research to fight against label noise.
Our insights further highlight that collecting the right PI in datasets requires some care to enable the learning of effective shortcuts.\looseness=-1

\section*{Acknowledgements}
We thank Jannik Kossen for helpful comments on this work. We also thank Josip Djolonga and Joan Puigcerver for helpful discussions related to infrastructure and data processing.

\bibliography{main}
\bibliographystyle{icml2023}

\newpage
\appendix
\onecolumn
\begin{appendices}

This appendix is organized as follows: In \cref{ap:imagenetpi} we describe the relabelling process done to generate ImageNet-PI. In \cref{ap:experimental_details} we describe in depth the experimental details of our experiments and our hyperparameter tuning strategy. \Cref{ap:low_noise} replicates our main findings in the low-noise version of the PI datasets. \Cref{ap:early_stopping} discusses and ablates the effect of early stopping in our experiments. \Cref{ap:other_datasets} provides additional results of our ablation studies on other datasets. \Cref{ap:trampp} and \cref{ap:afmpp} give further details about TRAM++ and AFM++, respectively. And, finally, \cref{ap:sop} and \cref{ap:het} describe in depth the experimental setup used to combine SOP and HET with TRAM, respectively.

\section{ImageNet-PI}\label{ap:imagenetpi}

\emph{ImageNet-PI} is a re-labelled version of the standard ILSVRC2012 ImageNet dataset in which the labels are provided by a collection of 16 deep neural networks with different architectures pre-trained on the standard ILSVRC2012. Specifically, the pre-trained models are downloaded from {\tt tf.keras.applications}\footnote{{\tt https://www.tensorflow.org/api\_docs/python/tf/keras/applications}}
and consist of: \texttt{
ResNet50V2, ResNet101V2, ResNet152V2, DenseNet121, DenseNet169, DenseNet201, InceptionResNetV2, InceptionV3, MobileNet, MobileNetV2, MobileNetV3Large, MobileNetV3Small, NASNetMobile, VGG16, VGG19, Xception}.

\begin{figure}[ht]
    \centering
    \includegraphics[width=0.3\textwidth]{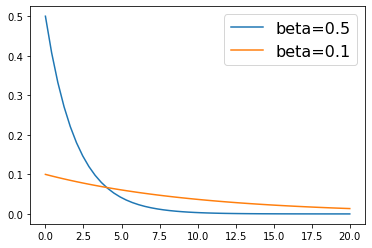}
    \caption{The effect of parameter $\beta$ in sampling temperatures.}
    \label{fig:beta_parameter}
\end{figure}

During the re-labelling process, we do not directly assign the maximum confidence prediction of each of the models, but instead, for each example, we sample a random label from the predictive distribution of each model on that example. Furthermore, to regulate the amount of label noise introduced when relabelling the dataset, ImageNet-PI allows the option to use  stochastic temperature-scaling to increase the entropy of the predictive distribution. The stochasticity of this process is controlled by a parameter $\beta$ which controls the inverse scale of a Gamma distribution (with shape parameter $\alpha=1.0$), from which the temperature values are sampled, with a code snippet looking as follows:
\begin{lstlisting}[language=Python]
# Get the predictive distribution of the model annotator.
pred_dist = model.predict(...)

# Sample the temperature.
temperature = tf.random.gamma(
    [tf.shape(pred_dist)[0]],
    alpha=tf.constant([1.]),
    beta=tf.constant([beta_parameter]))

# Compute the new predictive distribution.
log_probs = tf.math.log(pred_dist) / temperature
new_pred_dist = tf.nn.softmax(log_probs)

# Sample from the new predictive distribution.
class_predictions = tf.random.categorical(tf.math.log(new_pred_dist), 1)[:,0]
\end{lstlisting}
Intuitively, smaller values of $\beta$ translate to higher temperature values  as shown in \cref{fig:beta_parameter}, which leads to higher levels of label noise as softmax comes closer to uniform distribution for high temperatures.

This re-labelling process can produce arbitrarily noisy labels whose distribution is very far from being symmetrical, i.e., not all mis-classifications are equally likely. For example, it is more likely that similar dog breeds get confused among each other, but less likely that a `dog' gets re-labeled as a `chair'.

The PI in this dataset comes from the confidences of the models on the sampled label, their parameter count, and their test accuracy on the clean test distribution. These PI features are a good proxy for the expected reliability of each of the models. In our dataset release, we will provide the following files:
\begin{itemize}
    \item \textbf{labels-train.csv, labels-validation.csv} These files contain the new (noisy) labels for the training and validation set respectively. The new labels are provided by the pre-trained annotator models. Each file provides the labels in CSV format:\looseness=-1
    \begin{lstlisting}[language=Python]
    <image_id>,<label_1>,<label_2>,...,<label_16>
    \end{lstlisting}
   \item \textbf{confidences-train.csv, confidences-validation.csv} These files contain the confidence of each annotator model in its annotation; both for the training set and the validation set respectively. Each file provides the confidences in CSV format:
   \begin{lstlisting}[language=Python]
   <image_id>,<confidence_1>,<confidence_2>,...,<confidence_16>
   \end{lstlisting}
   \item \textbf{annotator-features.csv} This file contains the annotator features (i.e., meta-data about the model annotators themselves) in CSV format (16 rows; one for each model annotator):
   \begin{lstlisting}[language=Python]
   <model_accuracy>,<number_of_model_parameters>
   \end{lstlisting}
\end{itemize}

In particular, we will provide two standardized sampled annotations obtained by applying the temperature sampling process discussed above: one with $\beta=0.1$ corresponding to high label noise and one with $\beta=0.5$ corresponding to low label noise.\looseness=-1

\section{Experimental details}\label{ap:experimental_details}

We build upon the implementations and hyperparameters from the open source Uncertainty Baselines codebase \citep{nado2021uncertainty}. All results in the paper are reported based on 5 random seeds.

\subsection{Dataset-specific training settings}\label{ap:training_setup}

\subsubsection{CIFAR}

All CIFAR models are trained using a SGD optimizer with $0.9$ Nestrov momentum for $90$ epochs with a batch size of $256$. We sweep over an initial learning rate of $\{0.01, 0.1\}$ with the learning rate decayed by a factor of 0.2 after 27, 54 and 72 epochs. We sweep over an L2 regularization parameter of $\{0.00001, 0.0001, 0.001\}$. Following \citet{nado2021uncertainty}, we use a Wide ResNet model architecture with model-width multiplier of $10$ and a model-depth of $28$.

Unless specified otherwise, for TRAM and AFM models, we set the PI tower width to be $1024$ as this was the default parameter in \citet{tram}. We use the same architecture for the Distillation (PI) teacher. This controls the size of the subnetwork which integrates the PI which is parameterized as a concatenation of the pre-processed PI (and then passed through a Dense + ReLU layer) and the representation of the non-PI inputs by the base Wide ResNet model, followed by a Dense + ReLU layer, a residual connection and finally a concatenation of the joint feature space with the non-PI representation. The number of units in the Dense layers is controlled by the ``PI tower width".

For distillation models we uniformly sample over a temperature interval of $[0.5, 10]$. For CIFAR-10N and CIFAR-100N we split the original training set into a training and a validation set; $98\%$ of the examples are used for training and the remaining $2\%$ used as a validation set. Due to the smaller size of the CIFAR-10H training set (which following \citet{tram} is actually the original CIFAR test set), $96\%$ of the original training set is used as a training set with the remaining 4\% used as a validation set.

For TRAM++ and where relevant for AFM, we search over a no-PI loss weight of $\{0.1, 0.5\}$, a PI tower width of $\{512, 1024, 2048, 4096\}$ and a random PI length of $\{8, 14, 28\}$. For heteroscedastic CIFAR models, we set the number of factors for the low-rank component of the heteroscedastic covariance matrix \citep{het} to be $3$ for CIFAR-10H and CIFAR-10N, and $6$ for CIFAR-100N and search over $\{0.25, 0.5, 0.75, 1.0, 1.25, 1.5, 2.0, 3.0, 5.0\}$ for the heteroscedastic temperature.

\subsubsection{ImageNet-PI}

ImageNet models are trained using a SGD optimizer with $0.9$ Nestrov momentum for $90$ epochs with a batch size of $128$. We set the initial learning rate of $0.05$ with the learning rate decayed by a factor of $0.1$ after $30$, $60$ and $80$ epochs. We sweep over an L2 regularization parameter of $\{0.00001, 0.0001\}$. We use a ResNet-50 model architecture.

For TRAM and AFM models by default we set the PI tower width to be $2048$, with the same parameterization of the PI tower as for the CIFAR models. For distillation models we set the distillation temperature to be $0.5$. We use $1\%$ of the original ImageNet training set as a validation set.

For TRAM++ and where relevant for AFM, we set the no-PI loss weight of to $0.5$ and use random PI length of $30$. For heteroscedastic models we set the number of factors for the low-rank component of the heteroscedastic covariance matrix to be $15$ and search over $\{0.75, 1.0, 1.5, 2.0, 3.0\}$ for the heteroscedastic temperature.

\subsection{Hyperparameter tuning strategy}\label{ap:tuning}

Unless otherwise indicated, we report the test-set accuracy at the hyperparameters determined by the $\arg\max$ of best validation set accuracy (where the number of epochs are considered to be part of the set to be maximized over). The validation set used has noisy labels generated by the same process as the training set. This implements a realistic and noisy hyperparameter search with early stopping that we believe most closely replicates what is possible in real-world scenarios where a clean validation set may be unavailable. However, other papers report test-set metrics determined by a hyperparameter sweep assuming the availability of a clean validation set and/or without early stopping, which can have a large impact on the reported test-set metrics (see \cref{ap:early_stopping} for results computed in this way).

\clearpage
\section{Results on low-noise settings}\label{ap:low_noise}

In the main text, we always reported results for the high-noise settings of each of the datasets. However, we now show that all our findings from \cref{tab:pi_types} also apply in the low-noise setting.

\begin{table*}[ht]
\centering
\caption{Test accuracy of several methods trained using different features as PI \textit{\textbf{on the low-noise versions of the datasets}} (baselines in \gray{gray} and \emph{italics} do not use PI). Here, \orig{Original} denotes the standard PI of the dataset, \ind{Indicator} a binary signal that separates clean from noisy examples, \labs{Labels} the one-hot encoded labels, and \nearopt{Near-optimal} a synthetic feature consisting on giving the annotator label to those examples that are miss-annotated and a zero-vector otherwise. \textbf{Bold numbers} represent significant maximum values across PI features where significance means $\text{p-value}<0.05$.}
\label{tab:low_noise_pi_types}
\begin{tabular}{@{}clcccccc@{}}
\toprule
\multicolumn{1}{l}{} & \multicolumn{1}{c}{} & \orig{Original}                     & \ind{Indicator}              & \labs{Labels}                       & \nearopt{Near-optimal}                \\ \midrule
& \emph{no-PI}          & \gray{$85.8\scriptstyle{\pm 0.3}$} & \gray{$85.8\scriptstyle{\pm 0.3}$} &  \gray{$85.8\scriptstyle{\pm 0.3}$} & \gray{$85.8\scriptstyle{\pm 0.3}$} \\
 CIFAR10-H           & \emph{Distillation (no-PI)}   & \gray{$82.7\scriptstyle{\pm 0.0}$} & \gray{$82.7\scriptstyle{\pm 0.0}$} &  \gray{$82.7\scriptstyle{\pm 0.0}$} & \gray{$82.7\scriptstyle{\pm 0.0}$} \\
 (uniform)           & TRAM          & \orig{$85.1\scriptstyle{\pm 0.2}$} & \ind{$86.3\scriptstyle{\pm 0.3}$} &  \labs{$47.3\scriptstyle{\pm 1.7}$} & \nearopt{$\bm{87.6\scriptstyle{\pm 0.1}}$} \\
                     & Approximate FM            & \orig{$85.9\scriptstyle{\pm 0.1}$}   & \ind{$86.8\scriptstyle{\pm 0.2}$}   &  \labs{$38.1\scriptstyle{\pm 0.5}$}   & \nearopt{$\bm{87.4\scriptstyle{\pm 0.1}}$}   \\
                     & Distillation (PI)       & \orig{$81.0\scriptstyle{\pm 0.0}$} & \ind{$\bm{83.2\scriptstyle{\pm 0.3}}$} &  \labs{$81.1\scriptstyle{\pm 0.3}$} & \nearopt{$\bm{83.4\scriptstyle{\pm 0.3}}$} \\ \midrule
                     & \emph{no-PI}          & \gray{$88.3\scriptstyle{\pm 0.4}$} & \gray{$88.3\scriptstyle{\pm 0.4}$} &  \gray{$88.3\scriptstyle{\pm 0.4}$} & \gray{$88.3\scriptstyle{\pm 0.4}$} \\
CIFAR10-N            & \emph{Distillation (no-PI)}   & \gray{$85.4\scriptstyle{\pm 0.0}$} & \gray{$85.4\scriptstyle{\pm 0.0}$} &  \gray{$85.4\scriptstyle{\pm 0.0}$} & \gray{$85.4\scriptstyle{\pm 0.0}$} \\
(uniform)            & TRAM          & \orig{$88.7\scriptstyle{\pm 0.6}$} & \ind{$92.2\scriptstyle{\pm 0.3}$} &  \labs{$54.2\scriptstyle{\pm 0.1}$} & \nearopt{$\bm{93.7\scriptstyle{\pm 0.1}}$} \\
                     & Approximate FM            & \orig{$88.6\scriptstyle{\pm 0.6}$}   & \ind{$93.0\scriptstyle{\pm 0.2}$}   &  \labs{$37.4\scriptstyle{\pm 1.7}$}   & \nearopt{$\bm{94.0\scriptstyle{\pm 0.1}}$}   \\
                     & Distillation (PI)       & \orig{$88.1\scriptstyle{\pm 0.0}$} & \ind{$\bm{88.8\scriptstyle{\pm 0.6}}$} &  \labs{$\bm{88.8\scriptstyle{\pm 0.7}}$} & \nearopt{$\bm{88.6\scriptstyle{\pm 0.4}}$} \\ \midrule
 & \emph{no-PI}          & \gray{$68.3\scriptstyle{\pm 0.3}$} & \gray{$68.3\scriptstyle{\pm 0.3}$} &  \gray{$68.3\scriptstyle{\pm 0.3}$} & \gray{$68.3\scriptstyle{\pm 0.3}$} \\
ImageNet-PI          & \emph{Distillation (no-PI)}   & \gray{$67.9\scriptstyle{\pm 0.3}$} & \gray{$67.9\scriptstyle{\pm 0.3}$} &  \gray{$67.9\scriptstyle{\pm 0.3}$} & \gray{$67.9\scriptstyle{\pm 0.3}$} \\
(low-noise)          & TRAM          & \orig{$\bm{69.7\scriptstyle{\pm 0.3}}$} & \ind{$\bm{69.8\scriptstyle{\pm 0.2}}$} &  \labs{$55.9\scriptstyle{\pm 0.5}$} & \nearopt{$66.8\scriptstyle{\pm 0.1}$} \\
                     & Approximate FM            & \orig{$\bm{70.6\scriptstyle{\pm 0.2}}$}   & \ind{$\bm{70.7\scriptstyle{\pm 0.3}}$}   &  \labs{$24.9\scriptstyle{\pm 13.6}$}   & \nearopt{$\bm{70.8\scriptstyle{\pm 0.2}}$}   \\
                     & Distillation (PI)       & \orig{$\bm{67.2\scriptstyle{\pm 0.2}}$} & \ind{$\bm{67.4\scriptstyle{\pm 0.2}}$} &  \labs{$\bm{52.5\scriptstyle{\pm 26.3}}$} & \nearopt{$\bm{67.2\scriptstyle{\pm 0.7}}$} \\ \bottomrule
\end{tabular}
\end{table*}

\clearpage
\section{Effect of early stopping}\label{ap:early_stopping}

As early stopping is one of the strongest baselines against label noise, in all our experiments we held out a small portion of the noisy training set and reported clean test accuracy at the epoch with the best validation accuracy. However, to make sure that our findings do not depend on the use of early stopping, or the amount of label noise in the validation set, we now present a reproduction of the results in \cref{tab:pi_types} 
when either disabling early stopping or using a clean validation set to perform early stopping and hyperparameter tuning.

\subsection{No early stopping}

\Cref{tab:pi_types_no_early} shows the results of our benchmark without using early stopping. In general, we observe that without early stopping most baselines perform significantly worse as they overfit more to the noisy labels. In this regard, since one of the main benefits of PI is that it prevents memorization of the noisy labels, we see that without early stopping the relative improvement of the PI techniques with respect to their no-PI baselines is much larger. 

\begin{table*}[ht]
\centering
\caption{Test accuracy of several methods trained using different features as PI \textit{\textbf{not using early stopping}} (baselines in \gray{gray} and \emph{italics} do not use PI). Here, \orig{Original} denotes the standard PI of the dataset, \ind{Indicator} a binary signal that separates clean from noisy examples, \labs{Labels} the one-hot encoded labels, and \nearopt{Near-optimal} a synthetic feature consisting on giving the annotator label to those examples that are miss-annotated and a zero-vector otherwise. \textbf{Bold numbers} represent significant maximum values across PI features where significance means $\text{p-value}<0.05$.}
\label{tab:pi_types_no_early}
\begin{tabular}{@{}clcccccc@{}}
\toprule
\multicolumn{1}{l}{} & \multicolumn{1}{c}{} & \orig{Original}                     & \ind{Indicator}              & \labs{Labels}                       & \nearopt{Near-optimal}                \\ \midrule
& \emph{no-PI}          & \gray{$42.6\scriptstyle{\pm 0.3}$} & \gray{$42.6\scriptstyle{\pm 0.3}$} &  \gray{$42.6\scriptstyle{\pm 0.3}$} & \gray{$42.6\scriptstyle{\pm 0.3}$} \\
 CIFAR-10H           & \emph{Distillation (no-PI)}   & \gray{$45.2\scriptstyle{\pm 0.1}$} & \gray{$45.2\scriptstyle{\pm 0.1}$} &  \gray{$45.2\scriptstyle{\pm 0.1}$} & \gray{$45.2\scriptstyle{\pm 0.1}$} \\
 (worst)           & TRAM          & \orig{$59.2\scriptstyle{\pm 0.2}$} & \ind{$46.5\scriptstyle{\pm 0.6}$} &  \labs{$39.9\scriptstyle{\pm 0.3}$} & \nearopt{$\bm{77.4\scriptstyle{\pm 0.1}}$} \\
                     & Approximate FM            & \orig{$61.2\scriptstyle{\pm 0.7}$}   & \ind{$39.3\scriptstyle{\pm 0.6}$}   &  \labs{$10.0\scriptstyle{\pm 0.0}$}   & \nearopt{$\bm{79.3\scriptstyle{\pm 0.3}}$}   \\
                     & Distillation (PI)       & \orig{$45.2\scriptstyle{\pm 0.0}$} & \ind{$\bm{46.3\scriptstyle{\pm 0.1}}$} &  \labs{$45.7\scriptstyle{\pm 0.1}$} & \nearopt{$\bm{46.4\scriptstyle{\pm 0.1}}$} \\ \midrule
                     & \emph{no-PI}          & \gray{$67.7\scriptstyle{\pm 0.6}$} & \gray{$67.7\scriptstyle{\pm 0.6}$} &  \gray{$67.7\scriptstyle{\pm 0.6}$} & \gray{$67.7\scriptstyle{\pm 0.6}$} \\
CIFAR-10N            & \emph{Distillation (no-PI)}   & \gray{$71.4\scriptstyle{\pm 0.2}$} & \gray{$71.4\scriptstyle{\pm 0.2}$} &  \gray{$71.4\scriptstyle{\pm 0.2}$} & \gray{$71.4\scriptstyle{\pm 0.2}$} \\
(worst)            & TRAM          & \orig{$67.0\scriptstyle{\pm 0.4}$} & \ind{$79.2\scriptstyle{\pm 0.9}$} &  \labs{$51.6\scriptstyle{\pm 0.2}$} & \nearopt{$\bm{91.9\scriptstyle{\pm 0.2}}$} \\
                     & Approximate FM            & \orig{$69.8\scriptstyle{\pm 0.5}$}   & \ind{$89.6\scriptstyle{\pm 0.1}$}   &  \labs{$10.0\scriptstyle{\pm 0.0}$}   & \nearopt{$\bm{92.3\scriptstyle{\pm 0.2}}$}   \\
                     & Distillation (PI)       & \orig{$71.1\scriptstyle{\pm 0.2}$} & \ind{$\bm{73.1\scriptstyle{\pm 0.3}}$} &  \labs{$70.9\scriptstyle{\pm 0.3}$} & \nearopt{$70.9\scriptstyle{\pm 0.2}$} \\ \midrule
                     & \emph{no-PI}          & \gray{$55.8\scriptstyle{\pm 0.2}$} & \gray{$55.8\scriptstyle{\pm 0.2}$} &  \gray{$55.8\scriptstyle{\pm 0.2}$} & \gray{$55.8\scriptstyle{\pm 0.2}$} \\
           & \emph{Distillation (no-PI)}   & \gray{$58.6\scriptstyle{\pm 0.1}$} & \gray{$58.6\scriptstyle{\pm 0.1}$} &  \gray{$58.6\scriptstyle{\pm 0.1}$} & \gray{$58.6\scriptstyle{\pm 0.1}$} \\
CIFAR-100N           & TRAM          & \orig{$56.4\scriptstyle{\pm 0.3}$} & \ind{$56.0\scriptstyle{\pm 0.4}$} &  \labs{$34.9\scriptstyle{\pm 0.2}$} & \nearopt{$\bm{67.5\scriptstyle{\pm 0.3}}$} \\
                     & Approximate FM            & \orig{$58.9\scriptstyle{\pm 0.3}$}   & \ind{$65.4\scriptstyle{\pm 0.3}$}   &  \labs{$4.1\scriptstyle{\pm 0.1}$}   & \nearopt{$\bm{70.8\scriptstyle{\pm 0.4}}$}   \\
                     & Distillation (PI)       & \orig{$58.9\scriptstyle{\pm 0.2}$} & \ind{$58.9\scriptstyle{\pm 0.3}$} &  \labs{$56.8\scriptstyle{\pm 0.2}$} & \nearopt{$\bm{60.4\scriptstyle{\pm 0.4}}$} \\ \midrule
                     & \emph{no-PI}          & \gray{$47.7\scriptstyle{\pm 0.8}$} & \gray{$47.7\scriptstyle{\pm 0.8}$} &  \gray{$47.7\scriptstyle{\pm 0.8}$} & \gray{$47.7\scriptstyle{\pm 0.8}$} \\
ImageNet-PI          & \emph{Distillation (no-PI)}   & \gray{$50.4\scriptstyle{\pm 0.8}$} & \gray{$50.4\scriptstyle{\pm 0.8}$} &  \gray{$50.4\scriptstyle{\pm 0.8}$} & \gray{$50.4\scriptstyle{\pm 0.8}$} \\
(high-noise)          & TRAM          & \orig{$53.3\scriptstyle{\pm 0.4}$} & \ind{$53.5\scriptstyle{\pm 0.4}$} &  \labs{$41.0\scriptstyle{\pm 0.8}$} & \nearopt{$\bm{56.5\scriptstyle{\pm 0.3}}$} \\
                     & Approximate FM            & \orig{$55.0\scriptstyle{\pm 0.4}$}   & \ind{$55.5\scriptstyle{\pm 0.3}$}   &  \labs{$0.4\scriptstyle{\pm 0.1}$}   & \nearopt{$\bm{58.3\scriptstyle{\pm 0.1}}$}   \\
                     & Distillation (PI)       & \orig{$\bm{50.9\scriptstyle{\pm 0.4}}$} & \ind{$\bm{50.6\scriptstyle{\pm 0.2}}$} &  \labs{$39.1\scriptstyle{\pm 5.1}$} & \nearopt{$18.0\scriptstyle{\pm 24.7}$} \\ \bottomrule
\end{tabular}
\end{table*} 

\clearpage
\subsection{Clean validation set}
Most of the datasets we studied have a significant amount of label noise in their training set. In this regard, the small validation set we hold out from the training set is also very noisy, which can affect the performance of early stopping and hyperparameter tuning. For this reason, we also provide results in \cref{tab:pi_types_clean_val} in which we use the clean labels from the validation set for hyperparameter tuning and early stopping. As we can see, most methods perform  better in this regime, although our main findings about how the PI properties affect performance are still valid.
\begin{table*}[ht]
\centering
\caption{Test accuracy of several methods trained using different features as PI \textit{\textbf{using a clean validation set to select the best hyperparameters}} (baselines in \gray{gray} and \emph{italics} do not use PI). Here, \orig{Original} denotes the standard PI of the dataset, \ind{Indicator} a binary signal that separates clean from noisy examples, \labs{Labels} the one-hot encoded labels, and \nearopt{Near-optimal} a synthetic feature consisting on giving the annotator label to those examples that are miss-annotated and a zero-vector otherwise. \textbf{Bold numbers} represent significant maximum values across PI features where significance means $\text{p-value}<0.05$.}
\label{tab:pi_types_clean_val}
\begin{tabular}{@{}clcccccc@{}}
\toprule
\multicolumn{1}{l}{} & \multicolumn{1}{c}{} & \orig{Original}                     & \ind{Indicator}              & \labs{Labels}                       & \nearopt{Near-optimal}                \\ \midrule
& \emph{no-PI}          & \gray{$53.2\scriptstyle{\pm 1.0}$} & \gray{$53.2\scriptstyle{\pm 1.0}$} &  \gray{$53.2\scriptstyle{\pm 1.0}$} & \gray{$53.2\scriptstyle{\pm 1.0}$} \\
 CIFAR-10H           & \emph{Distillation (no-PI)}   & \gray{$53.4\scriptstyle{\pm 1.0}$} & \gray{$53.4\scriptstyle{\pm 1.0}$} &  \gray{$53.4\scriptstyle{\pm 1.0}$} & \gray{$53.4\scriptstyle{\pm 1.0}$} \\
 (worst)           & TRAM          & \orig{$67.7\scriptstyle{\pm 0.1}$} & \ind{$64.9\scriptstyle{\pm 0.6}$} &  \labs{$39.7\scriptstyle{\pm 0.3}$} & \nearopt{$\bm{77.4\scriptstyle{\pm 0.1}}$} \\
                     & Approximate FM            & \orig{$70.6\scriptstyle{\pm 0.4}$}   & \ind{$66.7\scriptstyle{\pm 2.1}$}   &  \labs{$29.5\scriptstyle{\pm 0.5}$}   & \nearopt{$\bm{79.1\scriptstyle{\pm 0.2}}$}   \\
                     & Distillation (PI)       & \orig{$\bm{53.9\scriptstyle{\pm 0.4}}$} & \ind{$\bm{53.3\scriptstyle{\pm 0.6}}$} &  \labs{$\bm{53.4\scriptstyle{\pm 0.4}}$} & \nearopt{$51.6\scriptstyle{\pm 0.0}$} \\ \midrule
                     & \emph{no-PI}          & \gray{$81.4\scriptstyle{\pm 0.5}$} & \gray{$81.4\scriptstyle{\pm 0.5}$} &  \gray{$81.4\scriptstyle{\pm 0.5}$} & \gray{$81.4\scriptstyle{\pm 0.5}$} \\
CIFAR-10N            & \emph{Distillation (no-PI)}   & \gray{$82.9\scriptstyle{\pm 0.4}$} & \gray{$82.9\scriptstyle{\pm 0.4}$} &  \gray{$82.9\scriptstyle{\pm 0.4}$} & \gray{$82.9\scriptstyle{\pm 0.4}$} \\
(worst)            & TRAM          & \orig{$81.9\scriptstyle{\pm 0.3}$} & \ind{$89.1\scriptstyle{\pm 0.3}$} &  \labs{$51.6\scriptstyle{\pm 0.1}$} & \nearopt{$\bm{91.1\scriptstyle{\pm 0.1}}$} \\
                     & Approximate FM            & \orig{$82.0\scriptstyle{\pm 0.3}$}   & \ind{$91.2\scriptstyle{\pm 0.3}$}   &  \labs{$22.6\scriptstyle{\pm 0.2}$}   & \nearopt{$\bm{92.3\scriptstyle{\pm 0.2}}$}   \\
                     & Distillation (PI)       & \orig{$\bm{80.8\scriptstyle{\pm 0.3}}$} & \ind{$\bm{80.7\scriptstyle{\pm 0.5}}$} &  \labs{$\bm{81.1\scriptstyle{\pm 0.4}}$} & \nearopt{$\bm{80.8\scriptstyle{\pm 0.2}}$} \\ \midrule
                     & \emph{no-PI}          & \gray{$60.8\scriptstyle{\pm 0.2}$} & \gray{$60.8\scriptstyle{\pm 0.2}$} &  \gray{$60.8\scriptstyle{\pm 0.2}$} & \gray{$60.8\scriptstyle{\pm 0.2}$} \\
           & \emph{Distillation (no-PI)}   & \gray{$60.8\scriptstyle{\pm 0.1}$} & \gray{$60.8\scriptstyle{\pm 0.1}$} &  \gray{$60.8\scriptstyle{\pm 0.1}$} & \gray{$60.8\scriptstyle{\pm 0.1}$} \\
CIFAR-100N            & TRAM          & \orig{$60.6\scriptstyle{\pm 0.3}$} & \ind{$63.3\scriptstyle{\pm 0.2}$} &  \labs{$34.8\scriptstyle{\pm 0.4}$} & \nearopt{$\bm{67.3\scriptstyle{\pm 0.3}}$} \\
                     & Approximate FM            & \orig{$60.2\scriptstyle{\pm 0.1}$}   & \ind{$67.8\scriptstyle{\pm 0.3}$}   &  \labs{$20.1\scriptstyle{\pm 0.3}$}   & \nearopt{$\bm{70.9\scriptstyle{\pm 0.2}}$}   \\
                     & Distillation (PI)       & \orig{$61.1\scriptstyle{\pm 0.2}$} & \ind{$\bm{61.9\scriptstyle{\pm 0.2}}$} &  \labs{$60.5\scriptstyle{\pm 0.2}$} & \nearopt{$61.5\scriptstyle{\pm 0.3}$} \\ \midrule
                     & \emph{no-PI}          & \gray{$48.3\scriptstyle{\pm 0.1}$} & \gray{$48.3\scriptstyle{\pm 0.1}$} &  \gray{$48.3\scriptstyle{\pm 0.1}$} & \gray{$48.3\scriptstyle{\pm 0.1}$} \\
ImageNet-PI          & \emph{Distillation (no-PI)}   & \gray{$50.5\scriptstyle{\pm 0.7}$} & \gray{$50.5\scriptstyle{\pm 0.7}$} &  \gray{$50.5\scriptstyle{\pm 0.7}$} & \gray{$50.5\scriptstyle{\pm 0.7}$} \\
(high-noise)          & TRAM          & \orig{$53.3\scriptstyle{\pm 0.3}$} & \ind{$53.8\scriptstyle{\pm 0.7}$} &  \labs{$40.7\scriptstyle{\pm 0.8}$} & \nearopt{$\bm{56.5\scriptstyle{\pm 0.2}}$} \\
                     & Approximate FM            & \orig{$55.6\scriptstyle{\pm 0.3}$}   & \ind{$55.5\scriptstyle{\pm 0.4}$}   &  \labs{$0.8\scriptstyle{\pm 0.2}$}   & \nearopt{$\bm{58.2\scriptstyle{\pm 0.1}}$}   \\
                     & Distillation (PI)       & \orig{$\bm{51.0\scriptstyle{\pm 0.4}}$} & \ind{$\bm{50.7\scriptstyle{\pm 0.3}}$} &  \labs{$39.1\scriptstyle{\pm 4.4}$} & \nearopt{$\bm{27.6\scriptstyle{\pm 22.7}}$} \\ \bottomrule
\end{tabular}
\end{table*}

\clearpage
\section{More results}\label{ap:other_datasets}

In this section, we provide complete results for the experiments in the main paper using other datasets and algorithms with the main findings.

\subsection{Training dynamics}\label{ap:other_dynamics}

\begin{figure*}[ht!]
    \centering
    \includegraphics[width=\textwidth]{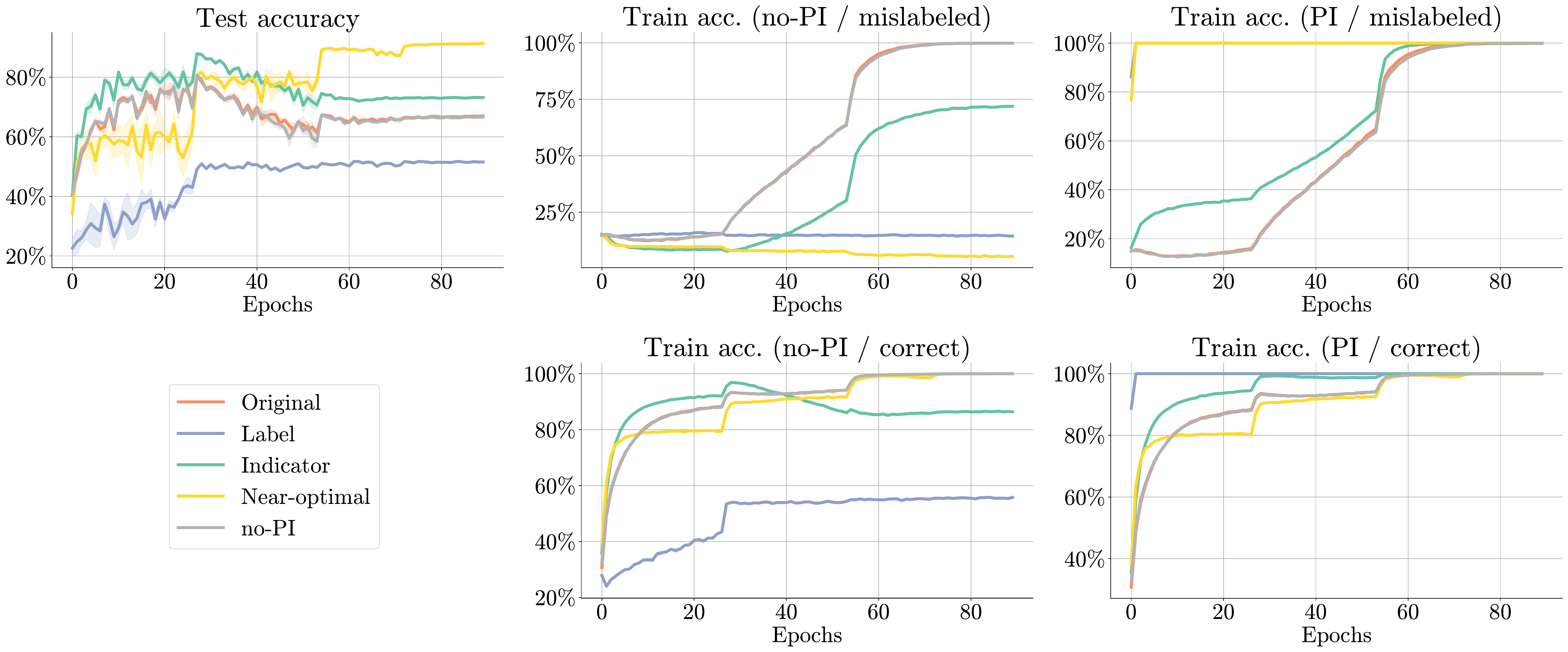}
    \caption{Dynamics of TRAM on CIFAR-10N with different PI features. \textbf{Top left}: Test accuracy. \textbf{Top center}: Train accuracy on noisy examples evaluated at the no-PI head, \textbf{Top right}: Train accuracy on noisy examples evaluated at the PI-head. \textbf{Bottom center}: Train accuracy on clean examples evaluated at the no-PI head. \textbf{Bottom right}: Train accuracy of clean examples evaluated at the PI head.}
    \label{fig:tram_dynamics_cifar10n}
    \vspace{-1em}
\end{figure*}

\begin{figure*}[ht!]
    \centering
    \includegraphics[width=\textwidth]{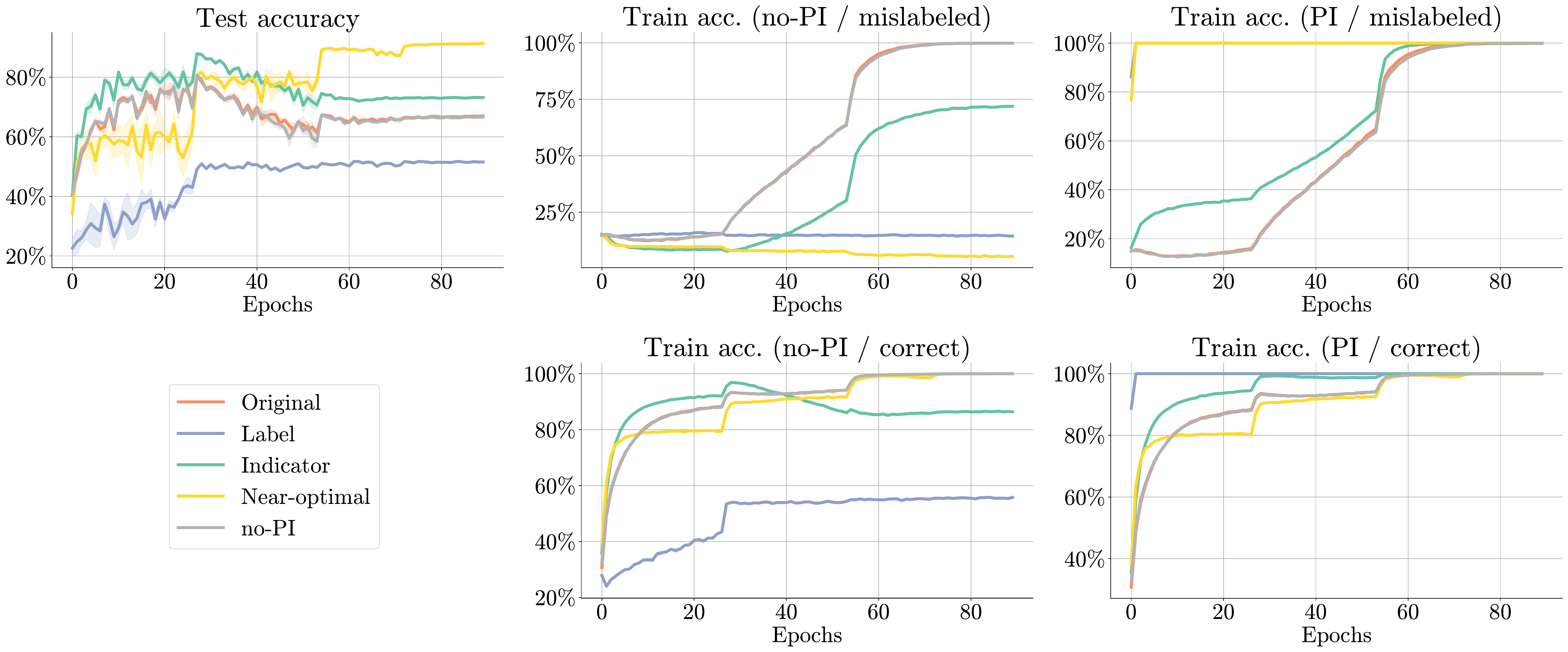}
    \vspace{-2em}
    \caption{Dynamics of TRAM on CIFAR-100N with different PI features. \textbf{Top left}: Test accuracy. \textbf{Top center}: Train accuracy on noisy examples evaluated at the no-PI head, \textbf{Top right}: Train accuracy on noisy examples evaluated at the PI-head. \textbf{Bottom center}: Train accuracy on clean examples evaluated at the no-PI head. \textbf{Bottom right}: Train accuracy of clean examples evaluated at the PI head.\looseness=-1}
    \label{fig:tram_dynamics_cifar100n}
    \vspace{-1em}
\end{figure*}

\begin{figure*}[ht!]
    \centering
    \includegraphics[width=\textwidth]{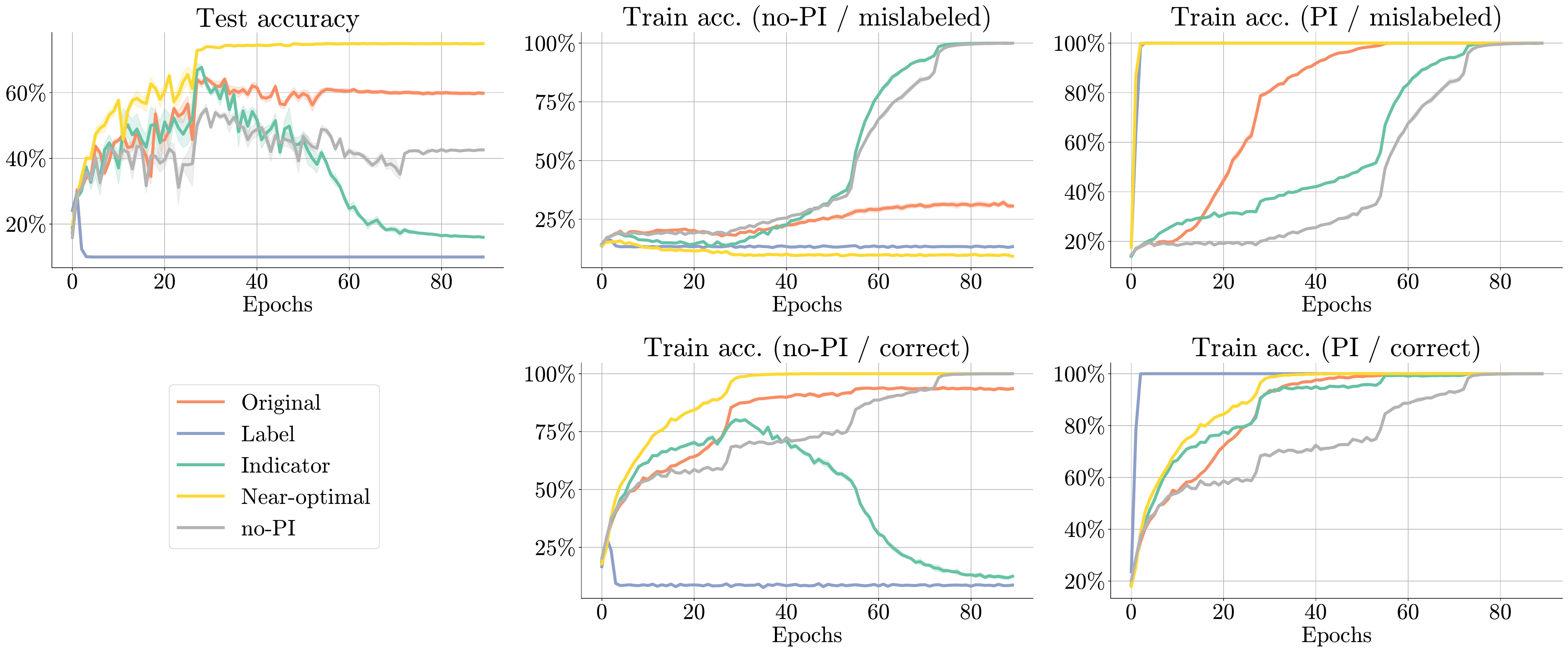}
    \caption{Dynamics of AFM on CIFAR-10H with different PI features. \textbf{Top left}: Test accuracy. \textbf{Top center}: Train accuracy on noisy examples evaluated with marginalization, \textbf{Top right}: Train accuracy on noisy examples evaluated at the PI-head. \textbf{Bottom center}: Train accuracy on clean examples evaluated with marginalization. \textbf{Bottom right}: Train accuracy of clean examples evaluated at the PI head.\looseness=-1}
    \label{fig:fm_dynamics_cifar10h}
    \vspace{-1em}
\end{figure*}

\begin{figure*}[ht!]
    \centering
    \includegraphics[width=\textwidth]{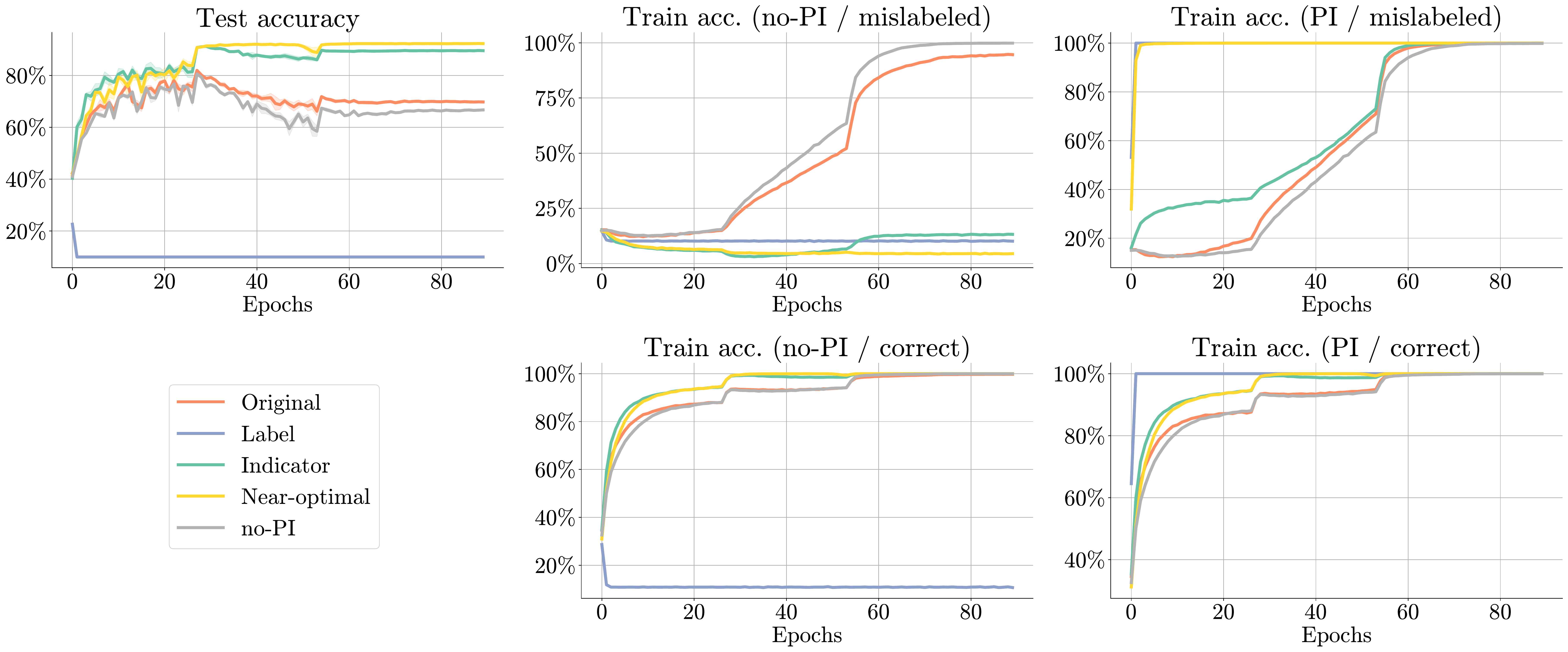}
    \caption{Dynamics of AFM on CIFAR-10N with different PI features. \textbf{Top left}: Test accuracy. \textbf{Top center}: Train accuracy on noisy examples evaluated with marginalization, \textbf{Top right}: Train accuracy on noisy examples evaluated at the PI-head. \textbf{Bottom center}: Train accuracy on clean examples evaluated with marginalization. \textbf{Bottom right}: Train accuracy of clean examples evaluated at the PI head.\looseness=-1}
    \label{fig:fm_dynamics_cifar10n}
    \vspace{-1em}
\end{figure*}

\begin{figure*}[ht!]
    \centering
    \includegraphics[width=\textwidth]{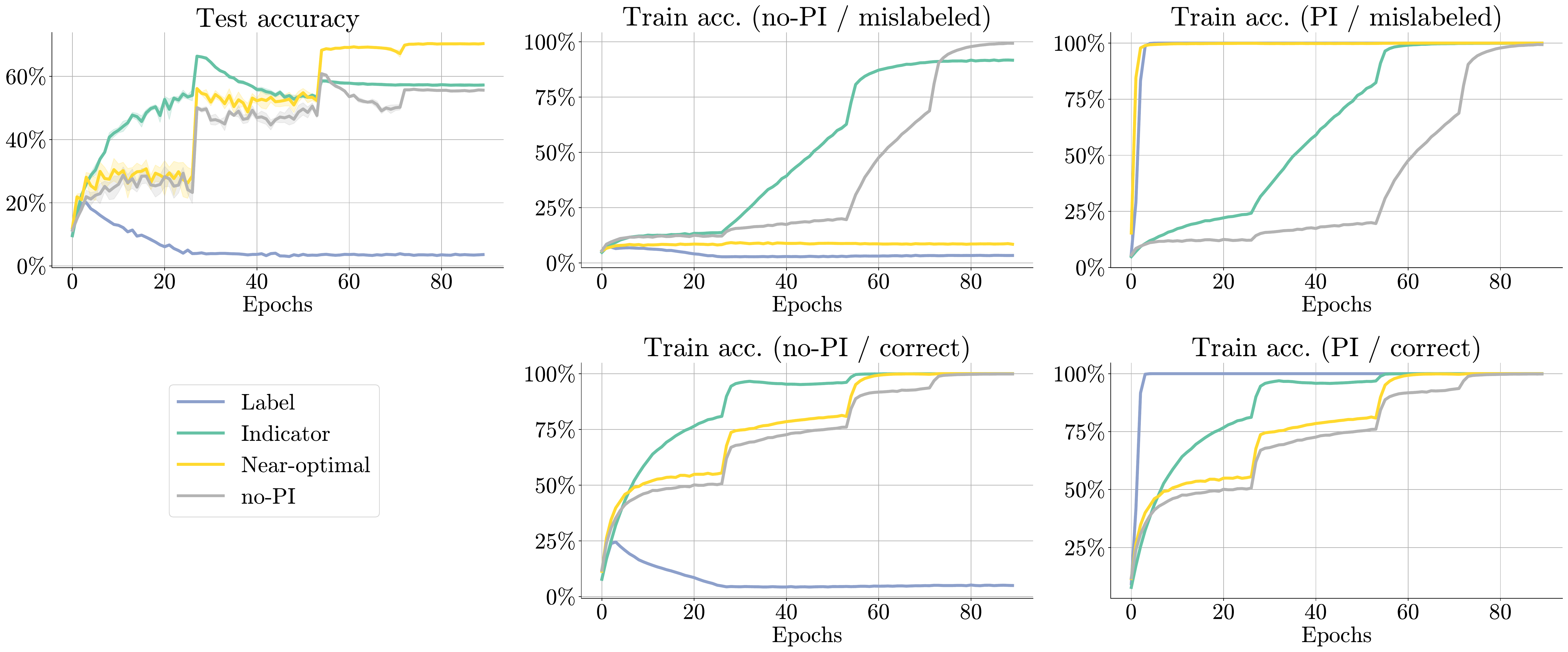}
    \caption{Dynamics of AFM on CIFAR-100N with different PI features. \textbf{Top left}: Test accuracy. \textbf{Top center}: Train accuracy on noisy examples evaluated with marginalization, \textbf{Top right}: Train accuracy on noisy examples evaluated at the PI-head. \textbf{Bottom center}: Train accuracy on clean examples evaluated with marginalization. \textbf{Bottom right}: Train accuracy of clean examples evaluated at the PI head.\looseness=-1}
    \label{fig:fm_dynamics_cifar100n}
    \vspace{-1em}
\end{figure*}

In \cref{fig:tram_dynamics} we provided a detailed analysis of the dynamics of TRAM on CIFAR-10H with different PI features. We now show results for TRAM on CIFAR-10N and CIFAR-100N (see \cref{fig:tram_dynamics_cifar10n} and \cref{fig:tram_dynamics_cifar100n}, respectively). We also show results for AFM on CIFAR-10H, CIFAR-10N, and CIFAR-100N (see \cref{fig:fm_dynamics_cifar10h},  \cref{fig:fm_dynamics_cifar10n} and \cref{fig:fm_dynamics_cifar100n}, respectively). 

\subsection{Feature extractor size}
\begin{figure}[ht!]
    \centering
    \includegraphics[width=0.7\textwidth]{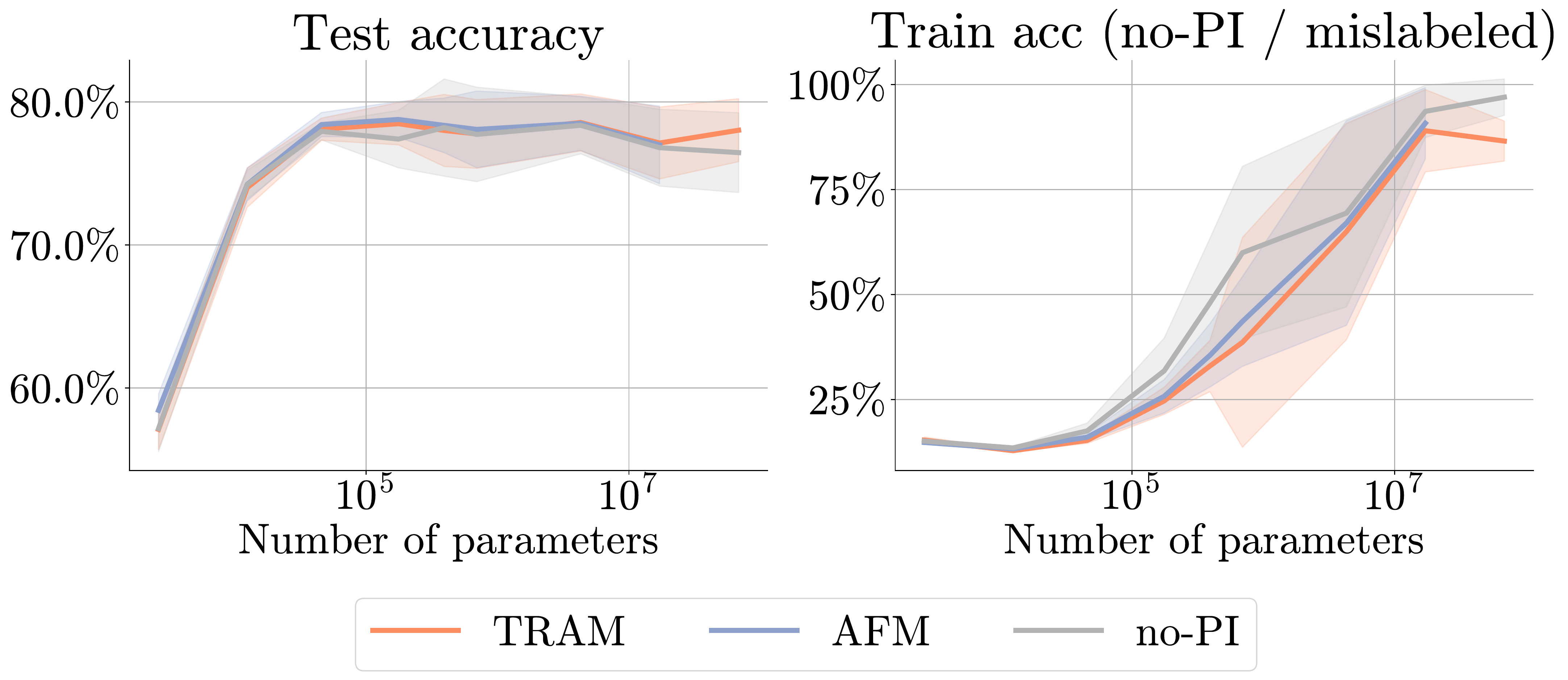}
    \vspace{-1em}
    \caption{Performance of different PI baselines on CIFAR-10N when increasing the parameter count of their feature extractor keeping the PI tower fixed. Larger models suffer from overfitting as they tend to use their larger capacity to overfit to noisy examples, discouraging the model from exploiting the PI.\looseness=-1} 
    \label{fig:feature_extractor_size_cifar10n}
    \vspace{-1em}
\end{figure}

\begin{figure}[ht!]
    \centering
    \includegraphics[width=0.7\textwidth]{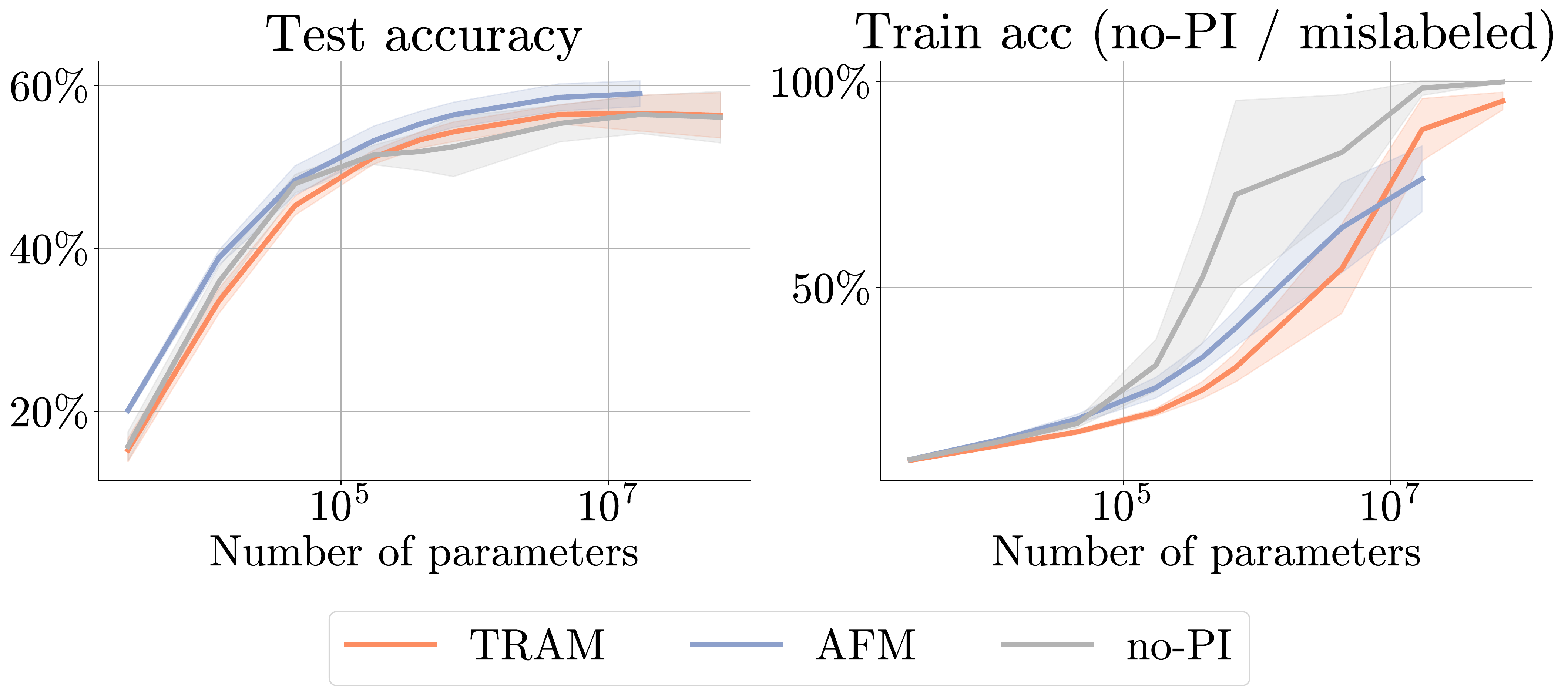}
    \vspace{-1em}
    \caption{Performance of different PI baselines on CIFAR-100N when increasing the parameter count of their feature extractor keeping the PI tower fixed. Larger models suffer from overfitting as they tend to use their larger capacity to overfit to noisy examples, discouraging the model from exploiting the PI.\looseness=-1} 
    \label{fig:feature_extractor_size_cifar100n}
    \vspace{-1em}
\end{figure}

We replicate the results in \cref{fig:feature_extractor_size} for other settings with the same findings. In particular, we show results on CIFAR-10N and CIFAR-100N (see \cref{fig:feature_extractor_size_cifar10n} and \cref{fig:feature_extractor_size_cifar100n}, respectively).

\subsection{PI head size}
\begin{figure}[ht]
    \centering
    \includegraphics[width=0.7\columnwidth]{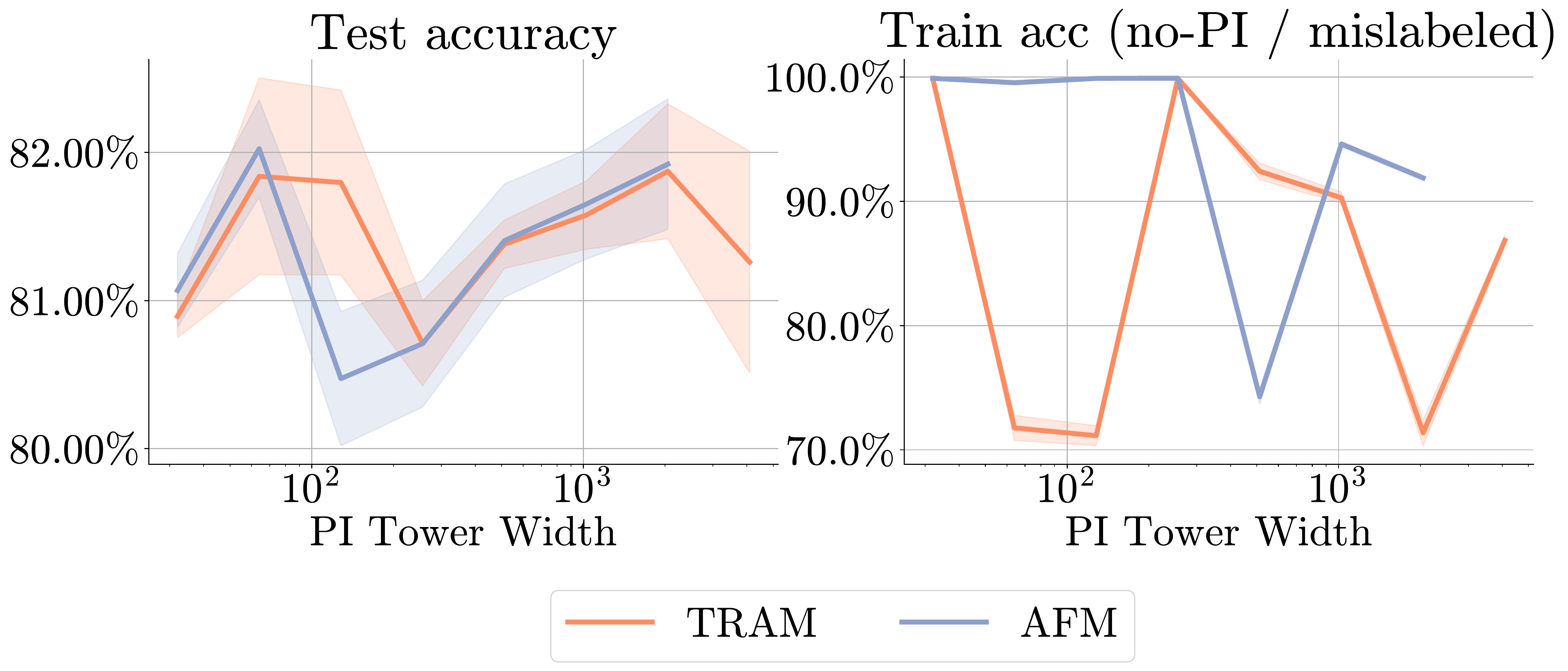}
    \vspace{-1em}
    \caption{Performance of different PI baselines on CIFAR-10N when increasing the PI head size. A larger PI head size incentivizes the model to memorize the noisy examples using the PI making more use of the PI as a shortcut.\looseness=-1}
    \label{fig:pi_head_size_cifar10n}
\end{figure}

\begin{figure}[ht]
    \centering
    \includegraphics[width=0.7\columnwidth]{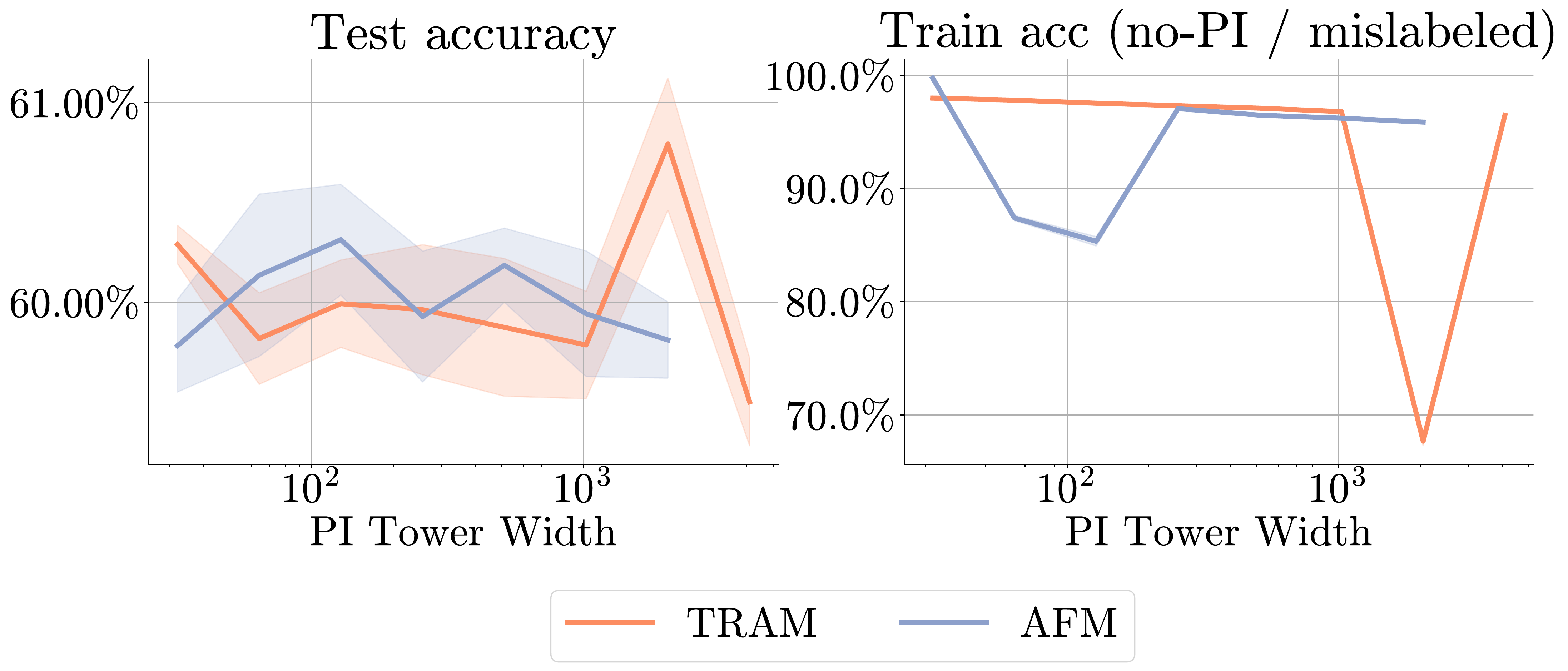}
    \vspace{-1em}
    \caption{Performance of different PI baselines on CIFAR-100N when increasing the PI head size. A larger PI head size incentivizes the model to memorize the noisy examples using the PI making more use of the PI as a shortcut.\looseness=-1}
    \label{fig:pi_head_size_cifar100n}    
\end{figure}

We replicate the results in \cref{fig:pi_head_size} on CIFAR-10N and CIFAR-100N (see \cref{fig:pi_head_size_cifar10n} and \cref{fig:pi_head_size_cifar100n}, respectively). In this case, however, we observe no clear trend in the results, probably due to the fact that the \orig{original} PI on these datasets is not good enough for TRAM and AFM to shine (cf. \cref{tab:pi_types}). In this regard, increasing the PI head size does not lead to better performance as there is nothing to extract from the PI. 

\clearpage
\section{Design details of TRAM++}\label{ap:trampp}

\begin{figure}[h]
    \centering
    \includegraphics[width=0.7\columnwidth]{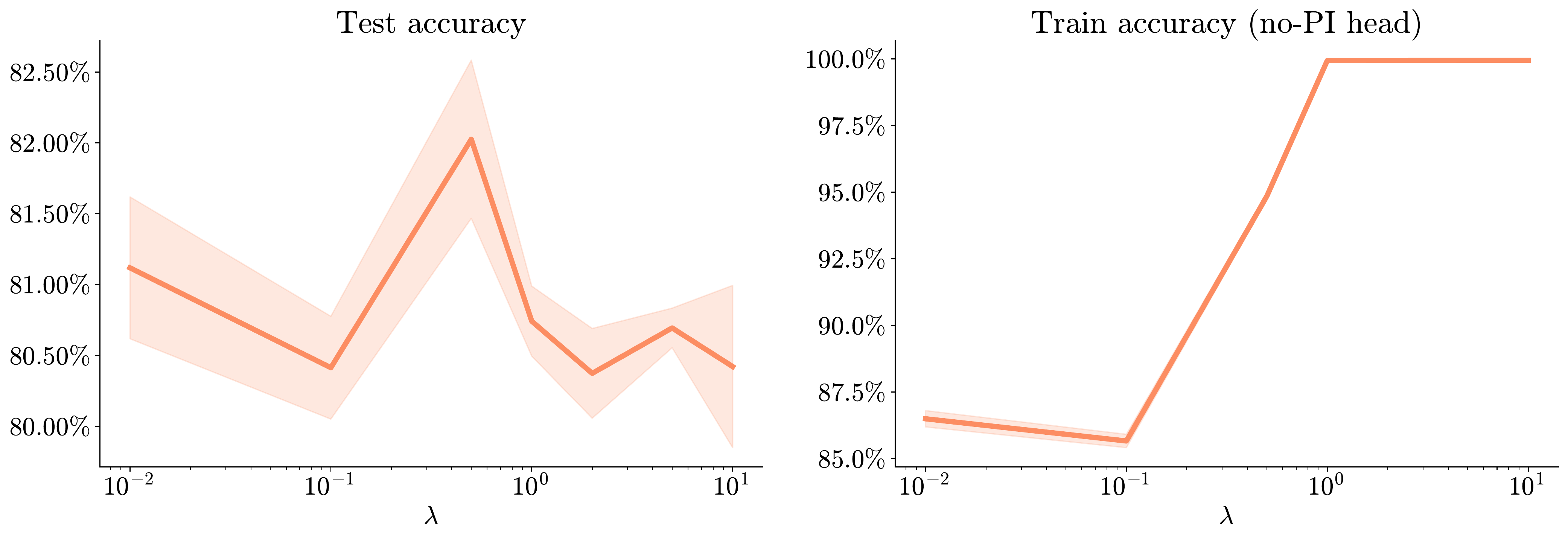}
    \caption{Performance of TRAM for different values of the loss weight $\lambda$ in CIFAR-10N. The optimal $\lambda$ is such the one that strikes a good balance between training the clean examples, while slowing down significantly the overfitting to the noisy ones.}
    \label{fig:lambda}
\end{figure}

We now give the design details for TRAM++, the improved version of TRAM which appends a unique random PI vector to the \orig{original} PI. In particular, we followed the same tuning strategy as in the rest of the TRAM experiments in the paper and we also tuned the parameter $\lambda$ that weighs the losses of the two heads, i.e.,
\begin{equation}
    \min_{\phi,\pi,\psi}\mathbb{E}_{(\bm x, \bm a, \tilde{y})}\left[\ell(\pi(\phi(\bm x), \bm a), \tilde{y})+\lambda\,\ell(\psi(\phi(\bm x)), \tilde{y})\right].
\end{equation}
\citet{tram} suggested that the gradients of the no-PI head do not affect the updates of the feature extraction, and thus $\lambda$ could be folded directly into the tuning of the global learning rate of TRAM. However, in our experiments, we found that tuning $\lambda$ given a fixed number of epochs can lead to significant gains in performance, as it can slow down training of the no-PI head. As seen in \cref{fig:lambda}, increasing $\lambda$ has the same effect as increasing the learning rate of the no-PI head, and a sweet spot exists for values of $\lambda<1$ in which the no-PI head trains fast enough to fit the clean examples, but avoids learning all the noisy ones.\looseness=-1

In general, $\lambda$ was not tuned in any of the other experiments, in order to remain as close as possible to the original TRAM implementation. However, for the TRAM++ experiments, which aimed to achieve the best possible performance out of TRAM, $\lambda$ was tuned.\looseness=-1

\section{Design of AFM++}\label{ap:afmpp}
In \cref{sec:random_pi}, we have seen that appending random PI that uniquely identifies each example to the \orig{original} PI can sometimes induce beneficial shortcuts in TRAM++. We now test the same strategy applied to AFM, and design AFM++, an augmented version of AFM with additional random PI. \Cref{tab:afm_improvements} shows the results of our experiments where we see that AFM++ also clearly improves over ``vanilla" AFM. Again, the improvements are greater in those datasets where overfitting is a bigger issue in the first place.

\begin{table*}[ht]
\centering
\caption{Performance comparison of no-PI, AFM and AFM++ on the different PI datasets.}
\label{tab:afm_improvements}
\begin{tabular}{@{}lccc@{}}
\toprule
                          & no-PI & AFM & AFM++ \\ \midrule
CIFAR-10H (worst)         & $55.0\scriptstyle{\pm 1.5}$ & $64.0\scriptstyle{\pm 0.6}$  & $\bm{68.2\scriptstyle{\pm 0.6}}$  \\
CIFAR-10N (worst)         & $80.6\scriptstyle{\pm 0.2}$ & $82.0\scriptstyle{\pm 0.3}$  & $\bm{84.6\scriptstyle{\pm 0.2}}$  \\
CIFAR-100N                & $60.4\scriptstyle{\pm 0.5}$ & $60.0\scriptstyle{\pm 0.2}$  & $\bm{61.9\scriptstyle{\pm 0.2}}$  \\
ImageNet-PI (high-noise)  & $47.7\scriptstyle{\pm 0.8}$ & $\bm{55.6\scriptstyle{\pm 0.3}}$  & $\bm{55.0\scriptstyle{\pm 0.6}}$ \\
\bottomrule
\end{tabular}
\end{table*}

\clearpage
\section{Combination of TRAM with label smoothing}\label{ap:label_smoothing}

We also evaluate the combination of TRAM with label smoothing (LS). In particular, we follow the standard label smoothing procedure and add the label smoothing hyperparameter to the hyperparameters swept over in \Cref{tab:pi_types}. More specifically, we sweep over label smoothing of $0.2$, $0.4$, $0.6$ and 0.8 and select the optimal hyperparameter setting following the same procedure as all experiments in the paper. The results are given in \Cref{tab:ls}.

We observe that on all datasets, adding label smoothing to the TRAM method leads to performance improvements, demonstrating that TRAM can be successfully combined with label smoothing. More generally, this observation strengthens the point that TRAM and TRAM++ are compatible and yield additive performance gains when combined with widely used methods developed for noisy labels.

\begin{table*}[ht]
\centering
\caption{Performance comparison of no-PI, Label smoothing (LS), TRAM TRAM + LS on different PI datasets.}
\label{tab:ls}
\begin{tabular}{@{}lcccc@{}}
\toprule
                          & no-PI & TRAM & LS & TRAM+LS \\ \midrule
CIFAR-10H (worst)         & $55.0\scriptstyle{\pm 1.5}$ & $64.9\scriptstyle{\pm 0.8}$ & $59.9\scriptstyle{\pm 1.5}$ & $\bm{65.4\scriptstyle{\pm 0.9}}$  \\
CIFAR-10N (worst)         & $80.6\scriptstyle{\pm 0.2}$ & $80.5\scriptstyle{\pm 0.5}$ & $80.5\scriptstyle{\pm 0.4}$ & $\bm{82.4\scriptstyle{\pm 0.2}}$  \\
CIFAR-100N                & $60.4\scriptstyle{\pm 0.5}$ & $59.7\scriptstyle{\pm 0.3}$ & $60.0\scriptstyle{\pm 0.46}$ & $\bm{61.9\scriptstyle{\pm 0.3}}$  \\
\bottomrule
\end{tabular}
\end{table*}

\section{Experimental details for SOP and TRAM+SOP}
\label{ap:sop}

As we have established in \cref{sec:other_techniques}, the combination of TRAM and SOP has the potential to achieve cumulative gains in robustness to label noise. TRAM, with its original PI, has been shown to improve performance on datasets with dense noise, such as CIFAR-10H (worst), compared to a model with no PI. However, the PI may not always be explanatory of the noise and even if it is, it may not fully explain away all of the noise. Additionally, the feature extractor and subsequent layers of the model may still be susceptible to noise, even when the PI is able to explain away the noise.

On the other hand, SOP has been shown to work well for sparsely distributed noise and operates on the principle of modeling out the noise, which is distinct from the method used by TRAM. As these principles are complementary to one another, we propose to combine the advantages of both methods to achieve cumulative gains.

As highlighted in \cref{sec:other_techniques}, the combination of TRAM+SOP consists of two main steps: pre-training with TRAM and fine-tuning with SOP. Our implementation of TRAM used regular TRAM with a few enhancements from TRAM++, such as random PI and a larger PI head size. It is important to note that our experiments were conducted using our own implementation of SOP and, although it incorporated the SOP method and was sanity-checked with the original authors of the paper, our experimental baseline environment and search space were different from theirs. As a result, the test accuracy on the CIFAR-N datasets may be lower than the results reported in the original SOP paper. However, the primary objective of these experiments was to explore whether TRAM + SOP can achieve cumulative gains over the respective implementations of TRAM and SOP alone and our results support this hypothesis.

In our experiments, both the SOP and TRAM+SOP models were trained for a total of 120 epochs, with a learning rate schedule that decayed at epochs 40, 80 and 110. We employed the SGD with Nesterov momentum for TRAM and regular momentum for SOP as in \citet{sop}. For a detailed description of the SOP parameters, we refer the reader to the original SOP paper. It is important to note that the results presented here for the TRAM+SOP method do not include all proposed enhancements in \citet{sop}. Further gains in performance may be achievable by incorporating these advancements and jointly optimizing the hyperparameter space for both the TRAM and SOP pretraining and fine-tuning stages.\looseness=-1

\section{Experimental details for TRAM+HET}\label{ap:het}

TRAM+HET consists of a simple two-headed TRAM model in which the last linear layer of each of the two heads has been substituted by a heteroscedastic linear layer~\citep{het}. In these experiments, we thus also sweep over the temperature of the heteroscedastic layers. A similar method was already proposed in \citet{tram}, under the name Het-TRAM, but here we also make use of our insights and allow the model to make use of random PI on top of the \orig{original} PI features. Interestingly, contrary to what happened with TRAM+SOP, the addition of random PI, i.e., TRAM++, did not always yield performance improvements using TRAM+HET. Instead, depending on the dataset (see \cref{tab:tramhet}) we observe that the use of random PI can sometimes hurt the final performance of the models (e.g., as in CIFAR-10H). We conjecture this might be due to the TRAM+HET models using the random PI to memorize the clean labels as well. Understanding why this happens only when using heteroscedastic layers is an interesting avenue for future work.

\begin{table*}[ht]
\centering
\caption{Performance comparison of TRAM, TRAM++, HET, TRAM+HET (without additional random PI), and TRAM+HET (with additional random PI) on the different PI datasets.}
\label{tab:tramhet}
\begin{tabular}{@{}lccccc@{}}
\toprule
                          & TRAM & TRAM++ & HET & TRAM+HET (w/o random) & TRAM+HET (+random)        \\ \midrule
CIFAR-10H (worst)         & $64.9\scriptstyle{\pm 0.8}$ & $66.8\scriptstyle{\pm 0.3}$  & $50.8 \scriptstyle{\pm 1.4}$ & $\bm{67.7 \scriptstyle{\pm 0.7}}$ &  $56.5 \scriptstyle{\pm 0.7}$ \\
CIFAR-10N (worst)         & $80.5\scriptstyle{\pm 0.5}$ & $83.9\scriptstyle{\pm 0.2}$  & $81.9 \scriptstyle{\pm 0.4}$ & $82.0 \scriptstyle{\pm 0.3}$ & $\bm{83.5\scriptstyle{\pm 0.1}}$  \\
CIFAR-100N         & $59.7\scriptstyle{\pm 0.3}$  & $61.1\scriptstyle{\pm 0.2}$ & $60.8 \scriptstyle{\pm 0.4}$ & $\bm{62.1\scriptstyle{\pm 0.1}}$  & $61.2 \scriptstyle{\pm 0.3}$ \\
ImageNet-PI (high-noise)  & $53.3 \scriptstyle{\pm 0.5}$  & $53.9 \scriptstyle{\pm 0.4}$ & $51.5 \scriptstyle{\pm 0.6}$ & $\bm{55.8 \scriptstyle{\pm 0.3}}$ & $\bm{55.4 \scriptstyle{\pm 0.4}}$ \\
\bottomrule
\end{tabular}
\end{table*}

\end{appendices}
\end{document}